%% file: main.tex
\documentclass{article}

\usepackage[final]{neurips_2022}

\usepackage{amsthm}
\usepackage{sidecap}
\usepackage{wrapfig}
\usepackage{bm}
\usepackage{amssymb}
\usepackage[utf8]{inputenc} 
\usepackage[T1]{fontenc}    
\usepackage{hyperref}       
\usepackage{url} 
\usepackage{bold-extra}
\usepackage{booktabs}       
\usepackage{amsfonts}       
\usepackage{nicefrac}       
\usepackage{microtype}      
\usepackage{amsmath}
\usepackage[numbers]{natbib}
\usepackage{graphicx}
\usepackage{multirow}
\usepackage{changepage,threeparttable} 
\usepackage{booktabs, multirow} 

\usepackage{soul}
\usepackage{multirow}
\usepackage{adjustbox}
\usepackage[table,xcdraw]{xcolor}
\usepackage{todonotes}%
\title{{\em Public Wisdom Matters}! Discourse-Aware Hyperbolic Fourier Co-Attention for Social-Text Classification}

\author{%
  Karish Grover \\
  IIIT Delhi\\
  India\\
  \texttt{karish19471@iiitd.ac.in}
  \And
  S.M. Phaneendra Angara \\
  LinkedIn \\
  India\\
  \texttt{sangara@linkedin.com}
  \And
  Md. Shad Akhtar \\
  IIIT Delhi \\
  India \\
  \texttt{shad.akhtar@iiitd.ac.in} \\
  \And
  Tanmoy Chakraborty \\
  IIT Delhi \\
  India \\
  \texttt{tanchak@ee.iitd.ac.in}
  }
  
\newcommand{\modelName}{{\ttfamily Hyphen}}
\begin{document}
\maketitle
\input{sections/abstract}
\input{sections/introduction}
\input{sections/related_work}
\input{sections/background}
\input{sections/methodology}
\input{sections/experiments}
\input{sections/conclusion}


\bibliographystyle{unsrt}
\bibliography{references}






\section*{Checklist}

\begin{enumerate}

\item For all authors...
\begin{enumerate}
  \item Do the main claims made in the abstract and introduction accurately reflect the paper's contributions and scope?
    \answerYes{}
  \item Did you describe the limitations of your work?
    \answerYes{} See Section \ref{sec:conclusion}
  \item Did you discuss any potential negative societal impacts of your work?
    \answerNA{}
  \item Have you read the ethics review guidelines and ensured that your paper conforms to them?
    \answerYes{}
\end{enumerate}

\item If you are including theoretical results...
\begin{enumerate}
  \item Did you state the full set of assumptions of all theoretical results?
    \answerYes{}
        \item Did you include complete proofs of all theoretical results?
    \answerNA{}
\end{enumerate}

\item If you ran experiments...
\begin{enumerate}
  \item Did you include the code, data, and instructions needed to reproduce the main experimental results (either in the supplemental material or as a URL)?
    \answerYes{} We release the Code and Data used as a part of the supplementary material.
  \item Did you specify all the training details (e.g., data splits, hyperparameters, how they were chosen)?
    \answerYes{} See experimentation details in Section \ref{sec:experiments}
        \item Did you report error bars (e.g., with respect to the random seed after running experiments multiple times)?
    \answerYes{} See Section \ref{sec:exp}
        \item Did you include the total amount of compute and the type of resources used (e.g., type of GPUs, internal cluster, or cloud provider)?
    \answerYes{} See experimentation details in Section \ref{sec:experiments}
\end{enumerate}

\item If you are using existing assets (e.g., code, data, models) or curating/releasing new assets...
\begin{enumerate}
  \item If your work uses existing assets, did you cite the creators?
    \answerYes{}
  \item Did you mention the license of the assets?
    \answerNA{}
  \item Did you include any new assets either in the supplemental material or as a URL?
    \answerYes{}
  \item Did you discuss whether and how consent was obtained from people whose data you're using/curating?
    \answerNA{}{}
  \item Did you discuss whether the data you are using/curating contains personally identifiable information or offensive content?
    \answerNA{}
\end{enumerate}

\item If you used crowdsourcing or conducted research with human subjects...
\begin{enumerate}
  \item Did you include the full text of instructions given to participants and screenshots, if applicable?
    \answerNA{}
  \item Did you describe any potential participant risks, with links to Institutional Review Board (IRB) approvals, if applicable?
    \answerNA{}
  \item Did you include the estimated hourly wage paid to participants and the total amount spent on participant compensation?
    \answerNA{}
\end{enumerate}

\end{enumerate}


\input{sections/appendix}

\end{document}

%% file: sections/abstract.tex
\begin{abstract}
Social media has become the fulcrum of all forms of communication.  Classifying social texts such as fake news, rumour, sarcasm, etc. has gained significant attention. The surface-level signals expressed by a social-text itself  may not be adequate for such tasks; therefore, recent methods attempted to incorporate other intrinsic signals such as user behavior and the underlying graph structure. Oftentimes, the `public wisdom' expressed through the comments/replies to a social-text acts as a surrogate of crowd-sourced view and may provide us with complementary signals. State-of-the-art methods on social-text classification tend to ignore such a rich hierarchical signal. Here, we propose \textbf{\texttt{Hyphen}}, a discourse-aware hyperbolic spectral co-attention network. \modelName\ is a fusion of hyperbolic graph representation learning with a novel Fourier co-attention mechanism in an attempt to {\em generalise} the social-text classification tasks by incorporating  {\em public discourse}. We parse public discourse as an Abstract Meaning Representation (AMR) graph and use the powerful hyperbolic geometric representation to model graphs with hierarchical structure. Finally, we equip it with a novel Fourier co-attention mechanism to capture the correlation between the source post and public discourse. Extensive experiments on four different social-text classification tasks, namely detecting fake news, hate speech, rumour, and sarcasm, show that \texttt{Hyphen} generalises well, and achieves state-of-the-art results on ten benchmark datasets. We also employ a sentence-level fact-checked and annotated dataset to evaluate how \modelName\ is capable of producing {\em explanations} as analogous evidence to the final prediction. Code is available at: \url{https://github.com/LCS2-IIITD/Hyphen}.

\end{abstract}

%% file: sections/introduction.tex
\section{Introduction}
\begin{figure}[!t]
\centering
\includegraphics[scale=0.30]{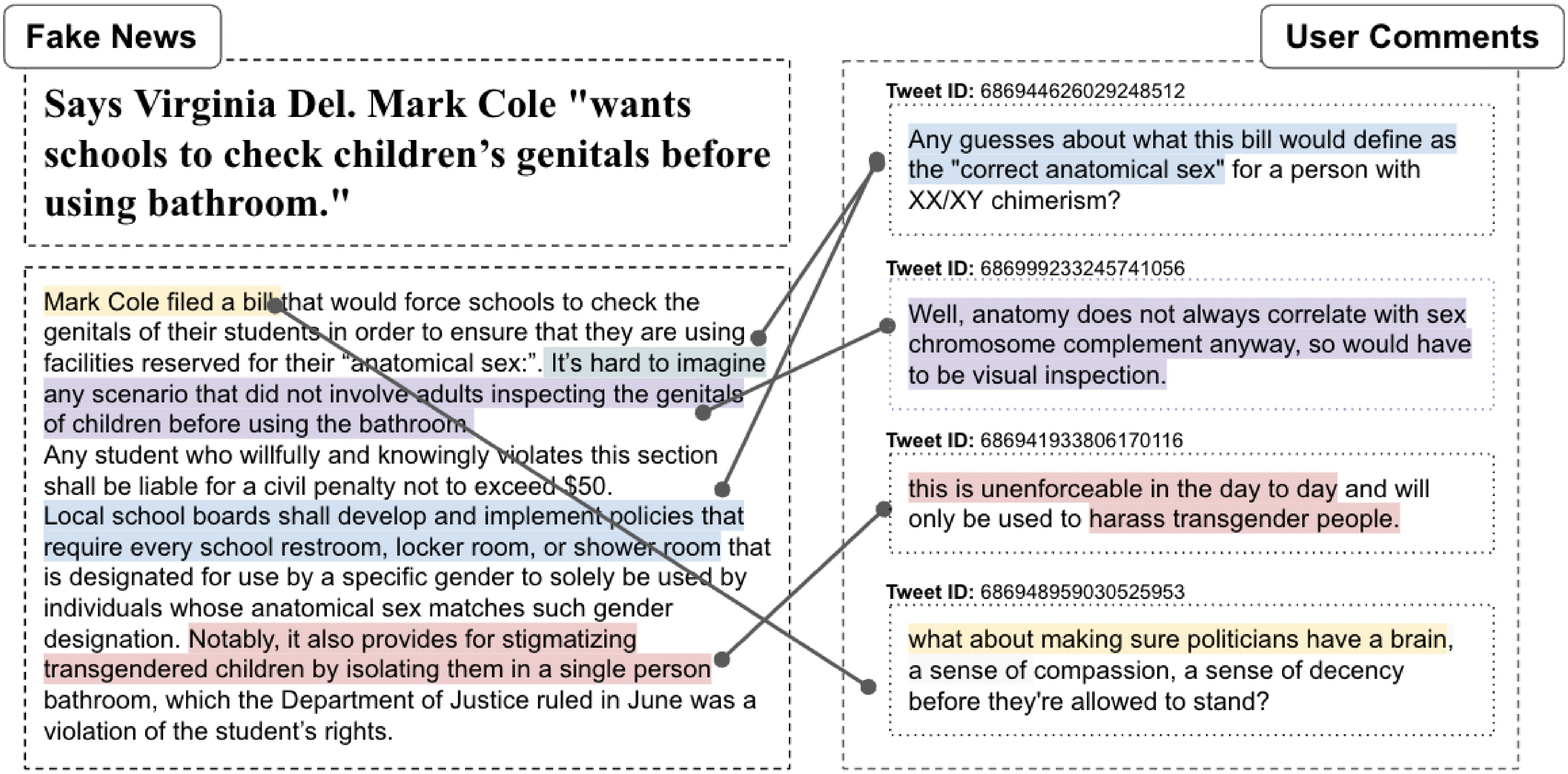}
\caption{A motivating example (taken from our dataset) showing how user comments act as analogous evidence for a fake news article. The third comment hints towards a possible sense of harassment being brought out by the highlighted portion of the text ({red}) and that it is a possible fake news. 
} \label{fig:teaser}
\vspace{-5mm}
\end{figure}

Social media has become a significant source of communication and information sharing. Mining texts shared on social media ({\em aka} social-texts) are indispensable for multiple tasks -- online offence detection, sarcasm identification, sentiment analysis, fake news detection, etc. Despite the proliferation of research in social computing, there is a gap in capturing the heterogeneous signals beyond the standalone source text processing. Predictive models incorporating signals such as user profiles \cite{unsvaag2018effects, shu2019role, malhotra2012studying, nurrahmi2018indonesian}, underlying user interaction networks \cite{yang2021rumor, guille2013predicting, pitas2016graph, guille2013information, lu2020gcan} and metadata information \cite{shu2019beyond, nguyen2020fang, zubiaga2017exploiting, gao2017detecting}, are far and few in between. These heterogeneous signals are challenging to obtain and may not always be available on different platforms (e.g., Reddit does not provide explicit user interaction network; YouTube does not  release user activities publicly).
On the other hand, comment threads following a source post are an equally rich source of heterogeneous signals, which are easier to obtain and uniformly available across social media platforms and forums. We hypothesise that such public discourse carries complementary and rich latent signals (public wisdom, worldly knowledge, fact busting, opinions, emotions, etc.), which would otherwise be difficult to obtain from just standalone source-post analysis. Therefore,  public discourse can be used in unison with the source posts to enhance social-text classification tasks. Figure \ref{fig:teaser} hints towards the motivation behind using public discourse as an implicit proxy for social-text classification.

In this work, we propose \modelName, a discourse-aware hyperbolic spectral co-attention
network that amalgamates the source post and its corresponding public discourse through a novel framework to perform generalised social-text classification. We parse individual comments on a source post as separate Abstract Meaning Representation (AMR) graphs \cite{banarescu2012abstract}, and merge them into one macro-AMR, representing mass perception and public wisdom about the source post. The AMR representation inherently abstracts away from syntactic representations, i.e., sentences which are similar in meaning are assigned similar AMRs, even if they are not identically worded \cite{flanigan2014discriminative}. The resultant macro-AMR graph represents the semantic information in a rich hierarchical structure \cite{https://doi.org/10.48550/arxiv.2110.07855, liu2018matching}. \modelName\ aims to effectively utilise the hierarchical properties of the macro-AMR graph by using the hyperbolic manifold \cite{cannon1997hyperbolic} for representation learning. 


In order to fuse the source post with the public discourse, we propose a novel Fourier co-attention mechanism on the hyperbolic space. It computes pair-wise attention between user comments and the source post, thereby capturing the correlation between them. On a typical social media post, several users express their opinions, and there are several messages being conveyed by the source post itself, some of which are more relevant and/or common than the others. We use a novel discrete Fourier transform based \cite{cooley1965algorithm} sublayer to filter the most-common user opinions expressed in the macro-AMR and most prominent messages being conveyed by the source post. Fourier transform is essentially a measurement of energy (i.e., strength of prevalence) of a particular frequency within a signal. We can extend this notion to quantify how dominant a particular frequency is within a signal. Building on this, we hypothesise that the time-domain signal isomorphically represents various user comments on the source post, and the Fourier transform over the comment representations yields the most-commonly occurring user \textit{frequencies} (stance, opinions, interpretation, wisdom, etc.). Similarly, the Fourier transform over the sentence-level representations of the source post renders the {\em most intense} messages and facts being conveyed by it. 

We perform extensive experiments with \modelName\ on four social-text classification tasks -- detecting fake news, hate speech, rumour, and sarcasm, on ten benchmark datasets. \modelName\ achieves state-of-the-art results across all datasets when compared with a suite of generic and data-specific baselines. Further, to evaluate the efficacy of hyperbolic manifold and Fourier co-attention in \modelName, we perform extensive ablation studies, which provide empirical justification behind the superiority of \modelName. Finally, we show how \modelName\ excels in producing explainability.  
 

%% file: sections/related_work.tex
\section{Related Work}

\textbf{Generic social-text classification.} There have been some attempts to arrive at a general architecture for social-text classification. Bi-RNODE \cite{https://doi.org/10.48550/arxiv.2112.12809} proposes to use recurrent neural ordinary differential equations by considering the time of posting. CBS-L \cite{fei2015social} considers transformation of document representation from the traditional $n$-gram feature space to a center-based similarity (CBS) space to solve the issue of co-variate shift. Pre-trained Transformer-based models like RoBERTa-base \cite{liu2019roberta}, BERTweet \cite{nguyen2020bertweet},  ClinicalBioBERT \cite{alsentzer2019publicly}, etc. also deliver benchmark results on generic social-text classification \cite{guo-etal-2020-benchmarking}. FNet \cite{https://doi.org/10.48550/arxiv.2105.03824} proves to be competent at modeling semantic relationships by replacing the self-attention layer in a Transformer encoder with a standard, non-parametric Fourier transform. 

\textbf{Use of public discourse in social-text classification.} Multiple approaches have been proposed to use public discourse as an attribute for classifying the social media posts. TCNN-URG \cite{qian2018neural} utilises a CNN-based network to encode the content, and a variational autoencoder for modeling user comments in fake news detection. CSI \cite{ruchansky2017csi} is a hybrid deep learning model that utilizes subtle clues from text, user responses, and the source post, while modeling the source post representation using an LSTM-based network. Zubiaga et al. \cite{zubiaga2018discourse} use public discourse for rumour stance detection using sequential classifiers. Lee et al. \cite{lee2018discourse} propose sentence-level distributed representation for the source post guided by the conversational structure. CASCADE \cite{https://doi.org/10.48550/arxiv.1805.06413} and CUE-CNN \cite{https://doi.org/10.48550/arxiv.1607.00976} use stylometric and personality traits of users in unison with the discussion threads to learn contextual representations for sarcasm detection. dEFEND \cite{shu2019defend} and GCAN \cite{lu2020gcan} propose to use co-attention over user comments and other social media attributes for detecting fake news and other social texts. The performance of most of these models deteriorate when extended to multiple tasks and fail to filter out the least relevant parts of their respective input modalities. Moreover, they operate on the Euclidean manifold, and therefore, overlook the representation strength of hyperbolic geometry in modeling hierarchical structures. \modelName\ overcomes these limitations of the existing methods.

\textbf{Hyperbolic representation learning.} Hyperbolic representation learning has gained significant attention in tasks in which the data inherently exhibits a hierarchical structure. HGCN \cite{chami2019hyperbolic} and HAT \cite{https://doi.org/10.48550/arxiv.1912.03046} achieve state-of-the-art results in graph classification owing to their powerful representation ability to model graphs with hierarchical structure. Unlike these two, H2H-GCN \cite{dai2021hyperbolic} directly works on the hyperbolic manifold to keep global hyperbolic structure, instead of relying on the tangent space. Furthermore, the recent GIL model \cite{https://doi.org/10.48550/arxiv.2010.12135} captures more informative internal structural features with low dimensions while maintaining conformal invariance of both Euclidean and hyperbolic spaces. However, for social-text classification, none of the above approaches simultaneously consider the source- and discourse-guided representations. We build on this limitation and use public comments in unison with the source post to further contextualise and improve a social-text classifier. 

%% file: sections/background.tex
\section{Background}
\textbf{Hyperbolic geometry.}
\label{sec:hyperbolic}
A Riemannian manifold $(\mathcal{M},g)$ of dimension $n$ is a real and smooth manifold equipped with an inner product on \textit{tangent} space $g_{\bm{x}}: \mathcal{T}_{\bm{x}}\mathcal{M} \times \mathcal{T}_{\bm{x}}\mathcal{M} \rightarrow \mathbb{R}$ at each point $\bm{x}\in \mathcal{M}$, where the \textit{tangent} space $\mathcal{T}_{\bm{x}}\mathcal{M}$ is an $n$-dimensional vector space and can be seen as a first-order local approximation of $\mathcal{M}$ around point $\bm{x}$.
In particular, hyperbolic space $(\mathbb{H}_c^n,g^c)$, a constant negative curvature Riemannian manifold, is defined by the manifold $\mathbb{H}_c^n=\{\bm{x}\in \mathbb{R}^n: c\lVert \bm{x}\rVert<1\}$ equipped with the following Riemannian metric:
\begin{math}
g_{\bm{x}}^{c}=\lambda_{\bm{x}}^2 g^E
\end{math}, where $\lambda_{\bm{x}}=\frac{2}{1-c\lVert \bm{x}\rVert^2}$, and $g^E=\bm{I}_n$ is the Euclidean metric tensor. The connections between hyperbolic space and \textit{tangent} space are established by the \textit{exponential} map $\operatorname{exp}_{\bm{x}}^c: \mathcal{T}_{\bm{x}}\mathbb{H}_c^n \rightarrow \mathbb{H}_c^n$, and the \textit{logarithmic} map $\operatorname{log}_{\bm{x}}^c: \mathbb{H}_c^n \rightarrow \mathcal{T}_{\bm{x}}\mathbb{H}_c^n$, as follows,
\begin{align}
\operatorname{exp}_{\bm{x}}^c(\bm{\mathbf{v}}) &= \bm{x}\oplus_c\left(\operatorname{tanh}\left(\sqrt{c}\frac{\lambda _{\bm{x}}^c \lVert \bm{\mathbf{v}}\rVert}{2}\right) \frac{\bm{\mathbf{v}}}{\sqrt{c}\lVert \bm{\mathbf{v}}\rVert}\right)\\
\operatorname{log}_{\bm{x}}^c(\bm{y}) &= \frac{2}{\sqrt{c}\lambda_{\bm{\mathbf{v}}}^c}\operatorname{tanh}^{-1}\left( \sqrt{c}\lVert -\bm{\mathbf{v}}\oplus_c \bm{\mathbf{v}}\rVert \right)\frac{-\bm{\mathbf{v}}\oplus_c \bm{\mathbf{v}}}{\lVert -\bm{x}\oplus_c \bm{\mathbf{v}} \rVert}
\end{align}

where $\bm{x},\bm{y}\in \mathbb{H}_c^n$, $\bm{v}\in \mathcal{T}_{\bm{x}}\mathbb{H}_c^n$, and $\oplus_c$ represents \textit{M$\ddot{\text{o}}$bius addition} as follows,
\begin{small}
\begin{equation}
\label{eq:mobius_add}
\bm{x}\oplus_c \bm{y} = \frac{(1+2c\langle \bm{x},\bm{y} \rangle+c\lVert \bm{y}\rVert^2)\bm{x}+(1-c\lVert \bm{x}\rVert^2)\bm{y}}{1+2c\langle \bm{x},\bm{y}\rangle + c^2\lVert \bm{x}\rVert^2\lVert \bm{y}\rVert^2}
\end{equation}
\end{small}Further, the generalization for multiplication in hyperbolic space can be defined by the \textit{M$\ddot{o}$bius matrix-vector multiplication} between vector $\bm{\mathbf{x}}\in \mathbb{H}_c^n \setminus\{\bm{0}\}$ and matrix $\mathbf{M}\in \mathbb{R}^{m\times n}$ as shown below,
\begin{small}
\begin{equation}
\label{eq:mobius_mul}
\mathbf{M} \otimes_c \bm{\mathbf{x}} = \frac{1}{\sqrt{c}} \tanh\left(
           \frac{\|\mathbf{M}\bm{\mathbf{x}}\|}{\|\bm{\mathbf{x}}\|}\tanh^{-1}(\sqrt{c}\|\bm{\mathbf{x}}\|)
        \right)\frac{\mathbf{M}\bm{\mathbf{x}}}{\|\mathbf{M}\bm{\mathbf{x}}\|}
\end{equation}
\end{small}

Hyperbolic space has been studied in differential geometry under five isometric models \cite{cannon1997hyperbolic}. This work mostly confines to the Poincaré ball model. It is a compact representation of the hyperbolic space and has the principled generalizations of basic operations (e.g., addition, multiplication). We use $\mathcal{P}, \mathcal{E}$ in the superscript, to denote the Poincaré and Euclidean manifolds, respectively. We provide more insights into these models in {\color{black}Appendix \ref{hyper-models}}. 

\textbf{Discrete Fourier Transform.} 
The Fourier transform decomposes a function into its constituent frequencies. Given a sequence $\{x_n\}$ with $n \in [0, N-1]$, the Discrete Fourier Transform (DFT) is defined as, $X_k = \sum_{n=0}^{N-1} x_n e^{-{\frac{2\pi i}{N}} n k}$, where $\quad 0 \leq k \leq N-1$. For each $k$, the DFT generates a new representation $X_k$ as a sum of the original input tokens $x_n$, with the \textit{twiddle factors} \cite{frigo2005design, cooley1965algorithm, https://doi.org/10.48550/arxiv.2111.13587}.

%% file: sections/methodology.tex
\section{Architecture of {\modelName}}
In this section, we lay out the structural details of \modelName\  (see Figure \ref{fig:model} for the schematic diagram). We propose individual pipelines for learning representations of the source post and the user comments on the hyperbolic space. We then combine both the representations using a novel hyperbolic Fourier co-attention mechanism that helps in simultaneously attending to both the representations. Lastly, we pass it to a feed-forward network for the final classification. Without loss of generality, we denote the Poincaré ball model ($\mathcal{P}$) as $\mathbb{H}_{c}^{n}$ (hyperbolic space) throughout the paper.

\subsection{Encoding public discourse}
In this section, we discuss the pipeline for encoding the public discourse. We parse the user comments into an AMR (Abstract Meaning Representation) \cite{banarescu2012abstract} graph. The individual comment-level AMR graphs are merged to form a macro-AMR (discussed below),  representing the global public wisdom and latent \textit{frequencies} in the discourse. Next, we learn representations of the macro-AMR using a HGCN (Hyperbolic Graph Convolutional Network) \cite{https://doi.org/10.48550/arxiv.1910.12933}. This yields a representation for public discourse containing rich latent signals.

\begin{figure}[!t]
\centering
\includegraphics[width = \textwidth]{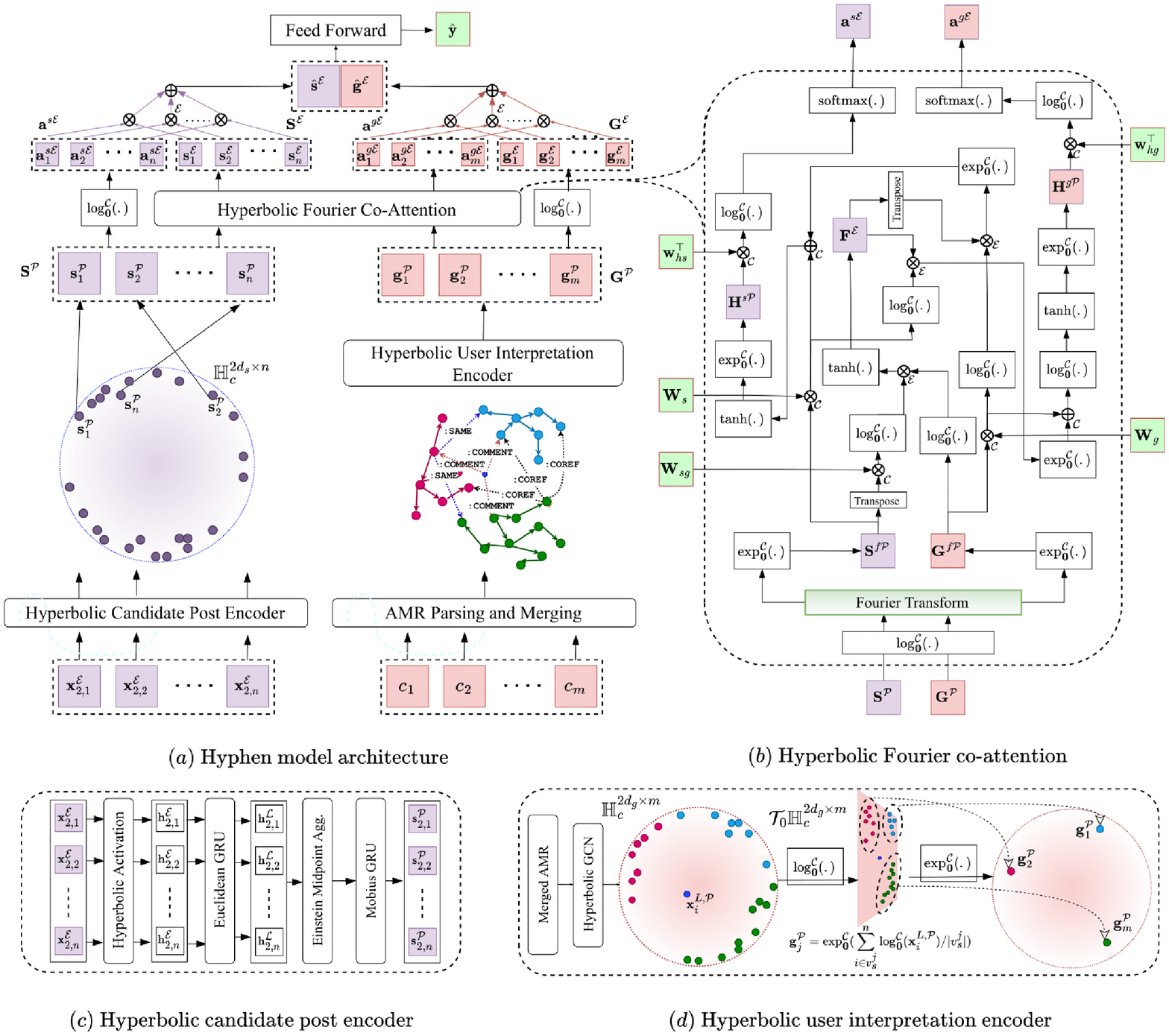}
\caption{Dissecting the primary components of \texttt{Hyphen}. The overall model architecture shown in (a) contains two parallel pipelines to encode the candidate post and user comments; (c) encodes the candidate post's sentences using an attention-enhanced hyperbolic word encoder (Section \ref{sec:content_encoder}), and (d) uses a hyperbolic GCN to encode the merged AMR containing latent user interpretations and form subgraph embeddings $\mathbf{g}_i^{\mathcal{P}}$ (Section \ref{sec:comment_encoder}). The final representations from (c) and (d), i.e., $\mathbf{S}^{\mathcal{P}}$ and $\mathbf{G}^{\mathcal{P}}$, are then passed to (b), which first transforms these through a Fourier sublayer and then computes the co-attention between user interpretation and the source post sentences in the hyperbolic space (Section \ref{sec:fourier}).}
\label{fig:model}
\end{figure}

\textbf{Macro-AMR graph creation}\label{sec:amr_merge}. Considering a social media post containing several user comments $C=[c_1,c_2,...,c_m]$, we obtain an AMR (Abstract Meaning Representation) \cite{banarescu2012abstract} graph for each user comment. We merge all the comment-level AMR graphs into one macro-AMR (post-level) while preserving the structural context of the subgraphs (comment-level). Figure \ref{fig:model}(a) contains the schematic for an example macro-AMR graph. In particular, we adopt three strategies -- (a) \textbf{\textit{Add a global dummy-node}}: We add a dummy node and connect it to all the root nodes of the comment-level AMRs, and add a comment tag \texttt{:COMMENT} to the edges. The dummy node ensures that all the AMRs are connected, so that information can be exchanged during graph encoding. (b) \textbf{\textit{Concept merging}}: Since we consider comments made on a particular post, these comments will essentially discuss the same topic. Therefore, multiple user comments can have identical mentions, resulting in repeated concept nodes in the comment-level AMRs. We identify such repeated concepts, and add an edge with label \texttt{:SAME} starting from \textit{earlier} nodes to \textit{later} nodes (here \textit{later} and \textit{earlier} refer to the temporal order of the ongoing conversation on a social media post). (c) \textbf{\textit{Inter-comment co-reference resolution}}: A major challenge for conversational understanding is posed by pronouns, which occur quite frequently in such social media comments. We conduct co-reference resolution on the comment-level AMRs to identify co-reference clusters containing concept nodes that refer to the same entity. We add edges labeled with the label \texttt{:COREF} between them, starting from \textit{earlier} nodes to \textit{later} nodes in a co-reference cluster to indicate their relation. Such types of connections can further enhance cross-comments information exchange. This step results in a post-level AMR graph $\mathcal{G}_{amr}=[g_s^1, g_s^2,...,g_s^m]$, representing relations between various subgraphs $\{g_s^i = (v_s^i, e_s^i) | 1 \le i \le m\}$ (different user comments will correspond to different subgraphs). The merged-AMR presents a global view of the public wisdom and interpretations.

\textbf{Hyperbolic graph encoder}\label{sec:comment_encoder}. We adopt the Poincaré ball model of  HGCN \cite{https://doi.org/10.48550/arxiv.1910.12933} to encode the post-level AMR graph and form user comment representations. Since different comments correspond to different subgraphs of the post-level AMR, we ultimately aggregate the node representations to form subgraph embeddings. Each  subgraph embedding represents how individual users interpret the source post (their opinion). In this section, we summarize the graph encoder architecture. Given a post-level AMR graph $\mathcal{G}_{amr}=(\mathcal{V}, E)$ and the Euclidean input node features, denoted by $(\mathbf{x}^{0,\mathcal{E}})  \in \mathbb{R}^{d_g}$, where $d_g$ is the input embedding dimension for entities in the AMR graph, we first map the input from Euclidean to Hyperbolic space. Therefore, we interpret $\mathbf{x}^{\mathcal{E}}$ as a point in the tangent space $\mathcal{T}_\mathbf{o}\mathbb{H}^{d_g}_c$ and map it to $\mathbb{H}^{d_g}_c$ with $\mathbf{x}^{0,\mathcal{P}}=\mathrm{exp}^c_\mathbf{0}(\mathbf{x}^{0,\mathcal{E}})$.
Our graph encoder then stacks multiple hyperbolic graph convolution layers to perform message passing (see \textcolor{black}{Appendix \ref{appendix:hgcn}} for the background on HGCN). Finally, we aggregate the hyperbolic node embeddings $(\mathbf{x}_i^{L,\mathcal{P}})_{i\in\mathcal{V}}$ at the last layer to form subgraph (comments) embeddings as shown in Figure \ref{fig:model}(d). We take the mean of the node embeddings for nodes present in a subgraph to yield the aggregated subgraph embedding: $\mathbf{g}^{\mathcal{P}}_j = \mathrm{exp}^c_\mathbf{0}(\sum_{i\in{v_s^j}}^{n}\mathrm{log}^c_\mathbf{0}(\mathbf{x}_i^{L,\mathcal{P}})/|v_s^j|)$. Here, $\mathbf{g}^{\mathcal{P}}_j \in \mathbb{H}^{2d_g}_c$, the operator $|.|$ represents the number of nodes present in subgraph $v_s^j$, and $L$ is the number of layers of HGCN. Therefore, the output for this encoder is  $\mathbf{G}^{\mathcal{P}}=[\mathbf{g_1}^{\mathcal{P}},\mathbf{g_2}^{\mathcal{P}},...,\mathbf{g_m}^{\mathcal{P}}]$, where $\mathbf{G}^{\mathcal{P}} \in \mathbb{H}^{2d_g\times m}$ is the matrix containing the learned representations for user interpretations (comments).

\subsection{Hyperbolic Candidate Post Encoder}\label{sec:content_encoder} 

Inspired by \cite{zhang2021hype}, we propose to learn the source post content representations through a hierarchical attention network in the hyperbolic space. We know that not all sentences in a source post might contain relevant information. We thus employ a hierarchical attention-based network to capture the relative importance of various sentences. Consider the input embedding of the $t^{th}$ word appearing in the $i^{th}$ sentence
as $\mathbf{x}_{it}$, in the candidate post. We utilise a hyperbolic word-level encoder (see \textcolor{black}{Appendix \ref{appendix:hhan}} for the background of Hyperbolic Hierarchical Attention Network (HHAN)) to learn $\mathbf{s}^{\mathcal{K}w}_i$, the representation of the $i^{th}$ sentence. Now, similar to the word-level encoder, we utilize \textit{M$\ddot{o}$bius}-GRU units to encode each sentence in the source post. We capture the sentence-level context to learn the sentence representation $\mathbf{s}^{\mathcal{P}}_i$ from the sentence vector $\mathbf{s}^{\mathcal{P}w}_i$ obtained from the word-level encoder. Specifically, we use Poincaré ball model based {\em M$\ddot{o}$bius}-GRU to encode different sentences. We obtain outputs from the \textit{M$\ddot{o}$bius}-GRU as $\mathbf{s}^{\mathcal{P}}_i = [\overrightarrow{GRU}_{mob}(\mathbf{s}^{\mathcal{P}w}_i), \overleftarrow{GRU}_{mob}(\mathbf{s}^{\mathcal{P}w}_i) ]$ as shown in Figure \ref{fig:model}(e). Here,  $\mathbf{s}^{\mathcal{P}}_i$ is the final context-aware representation for the $i^{th}$ sentence in the source post in the hyperbolic space (Poincaré ball model), i.e., $\mathbf{s}^{\mathcal{P}}_i \in \mathbb{H}^{2d_s}_c$, where $d_s$ is the input embedding dimension for the words $\mathbf{x}_{it}$ in the source post's sentences. This finally gives us $\mathbf{S}^{\mathcal{P}}=[\mathbf{s_1}^{\mathcal{P}},\mathbf{s_2}^{\mathcal{P}},...,\mathbf{s_n}^{\mathcal{P}}]$, where $\mathbf{S}^{\mathcal{P}} \in \mathbb{H}_c^{2d_s\times n}$ is the matrix containing the learned candidate post representations.

\subsection{Hyperbolic Fourier Co-Attention}
\label{sec:fourier}
We hypothesise that the evidence for various social-text classification tasks can be unveiled by investigating how different parts of the post are interpreted by different users, and how they correlate to different user opinions. Therefore, we develop a hyperbolic Fourier co-attention mechanism to model the mutual influence between the source social media post (i.e., $\mathbf{S}^{\mathcal{P}}=[\mathbf{s_1}^{\mathcal{P}},\mathbf{s_2}^{\mathcal{P}},...,\mathbf{s_n}^{\mathcal{P}}]$) and user comments (interpretation) embeddings (i.e., $\mathbf{G}^{\mathcal{P}}=[\mathbf{g_1}^{\mathcal{P}},\mathbf{g_2}^{\mathcal{P}},...,\mathbf{g_m}^{\mathcal{P}}]$, where $\mathbf{S}^{\mathcal{P}} \in \mathbb{H}_c^{2d_s\times n}$ and $\mathbf{G}^{\mathcal{P}} \in \mathbb{H}_c^{2d_g\times m}$). Co-Attention \cite{lu2016hierarchical} enables the learning of pairwise attentions, i.e., learning to attend based on computing word-level affinity scores between  two representations. Once we have the public discourse (Section \ref{sec:comment_encoder}) and the social media text (Section \ref{sec:content_encoder}) embeddings in the hyperbolic space, the next step is a Fourier sublayer, which applies a 2D DFT to its (sequence length, hidden dimension) embedding input -- one 1D DFT along the sequence dimension, $\mathcal{F}_{\textrm{seq}}$, and one 1D DFT along the hidden dimension, $\mathcal{F}_{\textrm{h}}$:\footnote{The relative ordering of $\mathcal{F}_{\textrm{seq}}$ and $\mathcal{F}_{\textrm{h}}$ in Equation \ref{eq:fourier_layer} is immaterial because the two 1D DFTs commute \cite{https://doi.org/10.48550/arxiv.2105.03824}.}
\begin{equation}
\label{eq:fourier_layer}
    \mathbf{S}^{f\mathcal{P}} = \mathrm{exp}^c_\mathbf{0}\left(\mathcal{F}_{\textrm{seq}}\left(\mathcal{F}_{\textrm{h}} \left( \mathrm{log}^c_\mathbf{0}(\mathbf{S}^{\mathcal{P}})\right)\right)\right),\,\, 
    \mathbf{G}^{f\mathcal{P}} = \mathrm{exp}^c_\mathbf{0}\left(\mathcal{F}_{\textrm{seq}}\left(\mathcal{F}_{\textrm{h}} \left( \mathrm{log}^c_\mathbf{0}(\mathbf{G}^{\mathcal{P}})\right)\right) \right)
\end{equation}

The intuition behind taking the Fourier transform over the user interpretation embeddings can be thought of as an attempt to capture the most commonly occurring \textit{frequencies} (public wisdom, worldly knowledge, fact busting, opinions, emotions, etc.) in the public discourse. These \textit{frequencies} signify how the source post is being received by most of the people. Further, the Fourier transform over the source post embeddings hints towards the most prominent messages conveyed by the source post. This is depicted in Figure \ref{fig:model}(b). Next, we compute a proximity matrix $\mathbf{F}^{\mathcal{E}}\in \mathbb{R}^{m \times n}$. The affinity (proximity) matrix $\mathbf{F}^{\mathcal{E}}$ can be thought to transform the user-interpretation attention space to the candidate post attention space, and vice versa for its transpose $\mathbf{F}^{\mathcal{E}\top}$. It is computed as:
\begin{equation}
    \mathbf{F}^{\mathcal{E}}=\text{tanh}\left(\mathrm{log}^c_\mathbf{0}\left(\mathbf{S}^{f\mathcal{P}\top}\otimes_c\mathbf{W}_{sg}\right)\otimes_{\mathcal{E}}\mathrm{log}^c_\mathbf{0}\left(\mathbf{G}^{f\mathcal{P}}\right)\right)
\end{equation}
where $\mathbf{W}_{sg} \in \mathbb{R}^{2d_s\times 2d_g}$ is a matrix of learnable parameters. The operator $\otimes_c$ is the \textit{M$\ddot{o}$bius Multiplication} operator (Equation \ref{eq:mobius_mul}), and $\otimes_{\mathcal{E}}$ is the simple euclidean matrix multiplication. By treating the affinity matrix as a feature, we can learn to predict candidate post and user interpretation attention maps $\mathbf{H}^{s\mathcal{P}} \in \mathbb{H}_c^{k \times n} \text{and } \mathbf{H}^{g\mathcal{P}} \in \mathbb{H}_c^{k \times m}$, given by
\begin{equation}
\begin{split}
\mathbf{H}^{s\mathcal{P}}&=\mathrm{exp}^c_\mathbf{0}(\text{tanh}(\mathrm{log}^c_\mathbf{0}(\mathbf{W}_s\otimes_c\mathbf{S}^{f\mathcal{P}}\oplus_c\mathrm{exp}^c_\mathbf{0}(\mathrm{log}^c_\mathbf{0}(\mathbf{W}_g\otimes_c\mathbf{G}^{f\mathcal{P}}))\otimes_{\mathcal{E}}\mathbf{F}^{\mathcal{E}\top}))))\\
\mathbf{H}^{g\mathcal{P}}&=\mathrm{exp}^c_\mathbf{0}(\text{tanh}(\mathrm{log}^c_\mathbf{0}(\mathbf{W}_g\otimes_c\mathbf{G}^{f\mathcal{P}}\oplus_c\mathrm{exp}^c_\mathbf{0}(\mathrm{log}^c_\mathbf{0}(\mathbf{W}_s\otimes_c\mathbf{S}^{f\mathcal{P}}))\otimes_{\mathcal{E}}\mathbf{F}^{\mathcal{E}}))))\\
\end{split}
\end{equation}
where $\mathbf{W}_s\in \mathbb{R}^{k\times 2d_s},\mathbf{W}_g\in \mathbb{R}^{k\times 2d_g}$ are learnable parameters, $k$ is the latent-dimension used in computing co-attention and $\oplus_c$ is the \textit{M$\ddot{o}$bius Addition} operator (Equation \ref{eq:mobius_add}). We can then generate the attention weights of source words and interaction users through the Softmax function:
\begin{equation}
\label{eq:attention_weights}
\mathbf{a}^{s\mathcal{E}} = \text{softmax}(\mathrm{log}^c_\mathbf{0}(\mathbf{w}_{hs}^{\top}\otimes_c\mathbf{H}^{s\mathcal{P}})), \quad \mathbf{a}^{g\mathcal{E}} = \text{softmax}(\mathrm{log}^c_\mathbf{0}(\mathbf{w}_{hg}^{\top}\otimes_c\mathbf{H}^{g\mathcal{P}}))
\end{equation}
where $\mathbf{a}^{s\mathcal{E}}\in \mathbb{R}^{1\times m}$ and $\mathbf{a}^{g\mathcal{E}}\in \mathbb{R}^{1\times n}$ are the vectors of attention probabilities for each sentence in the source story and each user comment, respectively. $\mathbf{w}_{hs}, \mathbf{w}_{hg} \in \mathbb{R}^{1\times k}$ are learnable weights. Eventually, we can generate the attention vectors of source sentences and user interpretation through weighted sum using the derived attention weights, given by
\begin{equation}
\hat{\mathbf{s}}^{\mathcal{E}}\\=\sum_{i=1}^{n}\mathbf{a}^{s\mathcal{E}}_i\mathbf{s}^{\mathcal{E}}_i~,~~~~\hat{\mathbf{g}}^{\mathcal{E}}=\sum_{j=1}^{m}\mathbf{a}^{g\mathcal{E}}_j\mathbf{g}^{\mathcal{E}}_j
\end{equation}
where $\hat{\mathbf{s}}^{\mathcal{E}}\in \mathbb{R}^{1\times 2d_s}$ and $\hat{\mathbf{g}}^{\mathcal{E}}\in \mathbb{R}^{1\times 2d_g}$ are the learned co-attention feature vectors that depict how sentences in the source post are correlated to the user interpretations. Finally, we have a feed forward network which yields the final classification output as $\hat{\mathbf{y}} = FFN[\hat{\mathbf{s}}^{\mathcal{E}}, \hat{\mathbf{g}}^{\mathcal{E}}]$, where $[.]$ is the concatenation operator. Equipped with co-attention learning, our model is further capable of generating suitable explanations (Section \ref{sec:exp}) by looking into the co-attention weights between different \textit{frequencies} of users and in the source post. 

%% file: sections/experiments.tex
\section{Experiments}
\label{sec:experiments}
\textbf{Datasets.} We evaluate the performance of \modelName\ on four different social-text classification tasks across ten datasets (c.f. Table \ref{tab:data_stats})  -- {(i)} fake news detection (Politifact \cite{shu2020fakenewsnet}, Gossipcop \cite{shu2020fakenewsnet}, AntiVax \cite{hayawi2022anti}), {(ii)} hate speech detection (HASOC \cite{mandl2019overview}), {(iii)} rumour detection (Pheme \cite{8118443}, Twitter15 \cite{ma-etal-2018-rumor}, Twitter16 \cite{ma-etal-2018-rumor}, RumourEval \cite{gorrell2019semeval}), and {(iv)} sarcasm detection (FigLang-Twitter\cite{ghosh2020report}, FigLang-Reddit \cite{ghosh2020report}). We augmented the datasets with public comments/replies to suite our experimental setting (see the {\color{black}Appendix \ref{appendix:experiments}} for details on dataset preparation). 

\begin{wraptable}{r}{10cm}
\tiny
\begin{tabular}{|
>{\columncolor[HTML]{FFFFFF}}c |
>{\columncolor[HTML]{FFFFFF}}c |
>{\columncolor[HTML]{FFFFFF}}c |
>{\columncolor[HTML]{FFFFFF}}c |
>{\columncolor[HTML]{FFFFFF}}c |
>{\columncolor[HTML]{FFFFFF}}c |}
\hline
\textbf{Dataset} & \textbf{\begin{tabular}[c]{@{}c@{}}\# source \\ posts\end{tabular}} & \textbf{\begin{tabular}[c]{@{}c@{}} Avg. \\ comments \\ (per post)\end{tabular}} & \textbf{\begin{tabular}[c]{@{}c@{}}SOTA-1\end{tabular}} & \textbf{\begin{tabular}[c]{@{}c@{}}SOTA-2\end{tabular}} & \textbf{\begin{tabular}[c]{@{}c@{}}SOTA-3\end{tabular}} \\ \hline
Politifact & 415 & 29 & *TCNN-URG \cite{qian2018neural} & HPA-BLSTM \cite{guo2018rumor} & *CSI \cite{ruchansky2017csi} \\ \hline
Gossipcop & 2813 & 20 & *TCNN-URG \cite{qian2018neural} & HPA-BLSTM \cite{guo2018rumor} & *CSI \cite{ruchansky2017csi} \\ \hline
Antivax & 3797 & 3 & *TCNN-URG \cite {qian2018neural} & HPA-BLSTM \cite {guo2018rumor} & *CSI \cite{ruchansky2017csi} \\ \hline
HASOC & 712 & 10 & CRNN & HPA-BLSTM & *CSI \cite{ruchansky2017csi} \\ \hline
Twitter15 & 543 & 9 & BiGCN \cite{chen2021bigcn} & GCAN \cite{lu2020gcan} & AARD \cite{song2021adversary} \\ \hline
Twitter16 & 362 & 27 & BiGCN \cite{chen2021bigcn} & GCAN \cite{lu2020gcan} & AARD \cite{song2021adversary} \\ \hline
Pheme & 6425 & 17 & DDGCN \cite{korban2020ddgcn} & *RumourGAN \cite{ma2019detect} & STS-NN \cite{huang2020deep} \\ \hline
Rumoureval$\dagger$ & 446 & 17 & CNN & DeClarE \cite{ye2021end} & \begin{tabular}[c]{@{}c@{}}MTL-\\ LSTM\cite{9498905}\end{tabular} \\ \hline
\begin{tabular}[c]{@{}c@{}}Figlang \\ Twitter\end{tabular} & 5000 & 4 & CNN + LSTM\cite{jain2020sarcasm} & \begin{tabular}[c]{@{}c@{}}Ensemble \{SVM, \\ LSTM, CNN-LSTM, \\ MLP\}\cite{lemmens-etal-2020-sarcasm}\end{tabular} & \begin{tabular}[c]{@{}c@{}}C-Net \cite{kumar-jena-etal-2020-c}\end{tabular} \\ \hline
\begin{tabular}[c]{@{}c@{}}Figlang \\ Reddit\end{tabular} & 4400 & 3 & CNN + LSTM \cite{jain2020sarcasm} & \begin{tabular}[c]{@{}c@{}}Ensemble \{SVM, \\ LSTM, CNN-LSTM, \\ MLP\} \cite{lemmens-etal-2020-sarcasm}\end{tabular} & \begin{tabular}[c]{@{}c@{}}C-Net \cite{kumar-jena-etal-2020-c}\end{tabular} \\ \hline
\end{tabular}
\caption{The statistics of the datasets and the chosen data-specific baselines for four social-text classification tasks. * denotes those baseline models which utilise public discourse. $\dagger$ denotes the dataset with three classes, and the remaining datasets have two levels.}
\label{tab:data_stats}
\end{wraptable}

\textbf{Experimentation details.} For both the hyperbolic encoders in Figures \ref{fig:model}(c)-(d), we adopt the Poincaré model of the respective frameworks. Due to limited machine precision, it is possible that the $\mathrm{exp}_0^c(.)$ and $\mathrm{log}_0^c(.)$ maps might sometimes return points that are not exactly located on the manifold. To avoid this and to ensure that points remain on the manifold and tangent vectors remain on the right tangent space, we clamp the maximum norm to $1 - e^{-14}$. For optimization on the hyperbolic space, we use Riemannian Adam from Geoopt \cite{https://doi.org/10.48550/arxiv.2005.02819}. To find the optimal $k$ (latent dimension, see Equation \ref{eq:fourier_layer}) for hyperbolic co-attention, we run grid search over $k = {50, 80, 128, 256}$, and finally use $k = 128$. For HGCN, we use two layers with curvatures $K_1$ = $K_2$ = -1. We detail all other hyper-parameters in the \textcolor{black}{Appendix \ref{appendix:experiments}}. We run all experiments for 100 epochs with early stopping patience of 10 epochs, on a NVIDIA RTX A6000 GPU. 
\begin{table}[]
\label{tab:comparison}
\resizebox{\columnwidth}{!}{%
\begin{tabular}{|p{0.1cm}|c|l|c|c| c |c|c | c| c | c| c|}
\hline
\multicolumn{1}{|c|}{\multirow{2}{*}{\textbf{Task}}} & \multicolumn{1}{|c|}{\multirow{2}{*}{\textbf{Dataset}}} & \multicolumn{1}{|c|}{\multirow{2}{*}{\textbf{}}} & \multicolumn{3}{c|}{\textbf{Data-specific baseline}} & \multicolumn{4}{c|}{\textbf{Generic neural baseline}} & \multicolumn{2}{c|}{\textbf{\modelName}} \\ \cline{4-12}
& & & \multicolumn{1}{c|}{\textbf{SOTA-1}} & \multicolumn{1}{c|}{\textbf{SOTA-2}} & \multicolumn{1}{c|}{\textbf{SOTA-3}} & \multicolumn{1}{c|}{\textbf{HAN}} & \multicolumn{1}{c|}{\textbf{dEFEND}} & \multicolumn{1}{c|}{\textbf{BERT}} & \multicolumn{1}{c|}{\textbf{RoBERTa}} & \multicolumn{1}{c|}{\textbf{Eucli.}} & \multicolumn{1}{c|}{\textbf{Hyper.}} \\ \hline
\multicolumn{1}{|c|}{} & \multicolumn{1}{c|}{} & Pre. & \multicolumn{1}{c|}{0.712} & \multicolumn{1}{c|}{0.894} & 0.847 & \multicolumn{1}{c|}{0.852} & \multicolumn{1}{c|}{0.902} & \multicolumn{1}{c|}{0.911} & \underline{ 0.924} & \multicolumn{1}{c|}{0.951} & \textbf{0.972} \\
\multicolumn{1}{|c|}{} & \multicolumn{1}{c|}{} & Rec. & \multicolumn{1}{c|}{0.785} & \multicolumn{1}{c|}{0.868} & 0.897 & \multicolumn{1}{c|}{\underline{0.958}} & \multicolumn{1}{c|}{0.956} & \multicolumn{1}{c|}{0.904} & 0.903 & \multicolumn{1}{c|}{0.936} & \multicolumn{1}{c|}{\textbf{0.961}} \\
\multicolumn{1}{|c|}{} & \multicolumn{1}{c|}{\multirow{-3}{*}{\textbf{Politifact}}} & F1 & \multicolumn{1}{c|}{0.827} & \multicolumn{1}{c|}{0.881} & 0.871 & \multicolumn{1}{c|}{0.902} & \multicolumn{1}{c|}{\underline{ 0.928}} & \multicolumn{1}{c|}{0.905} & 0.906 & \multicolumn{1}{c|}{0.940} & \multicolumn{1}{c|}{\textbf{0.968}} \\ \cline{2-12}
\multicolumn{1}{|c|}{} & \multicolumn{1}{c|}{} & Pre. & \multicolumn{1}{c|}{0.715} & \multicolumn{1}{c|}{0.684} & 0.732 & \multicolumn{1}{c|}{\textbf{0.818}} & \multicolumn{1}{c|}{0.729} & \multicolumn{1}{c|}{0.764} & 0.771 & \multicolumn{1}{c|}{0.786} & \multicolumn{1}{c|}{\underline{ 0.791}} \\
\multicolumn{1}{|c|}{} & \multicolumn{1}{c|}{} & Rec. & \multicolumn{1}{c|}{0.521} & \multicolumn{1}{c|}{0.662} & 0.638 & \multicolumn{1}{c|}{0.742} & \multicolumn{1}{c|}{\underline{ 0.782}} & \multicolumn{1}{c|}{0.761} & 0.775 & \multicolumn{1}{c|}{0.776} & \multicolumn{1}{c|}{\textbf{0.788}} \\
\multicolumn{1}{|c|}{} & \multicolumn{1}{c|}{\multirow{-3}{*}{\textbf{Gossipcop}}} & F1 & \multicolumn{1}{c|}{0.603} & \multicolumn{1}{c|}{0.673} & 0.682 & \multicolumn{1}{c|}{\underline{ 0.778}} & \multicolumn{1}{c|}{0.755} & \multicolumn{1}{c|}{0.762} & 0.772 & \multicolumn{1}{c|}{0.781} & \multicolumn{1}{c|}{\textbf{0.816}} \\ \cline{2-12} 
\multicolumn{1}{|c|}{} & \multicolumn{1}{c|}{} & Pre. & \multicolumn{1}{c|}{0.829} & \multicolumn{1}{c|}{0.865} & 0.901 & \multicolumn{1}{c|}{0.806} & \multicolumn{1}{c|}{0.935} & \multicolumn{1}{c|}{0.943} & \underline{ 0.948} & \multicolumn{1}{c|}{0.941} & \multicolumn{1}{c|}{\textbf{0.951}} \\
\multicolumn{1}{|c|}{} & \multicolumn{1}{c|}{} & Rec. & \multicolumn{1}{c|}{0.825} & \multicolumn{1}{c|}{0.864} & 0.912 & \multicolumn{1}{c|}{0.862} & \multicolumn{1}{c|}{0.934} & \multicolumn{1}{c|}{\underline{ 0.941}} & \textbf{0.961} & \multicolumn{1}{c|}{0.937} & \multicolumn{1}{c|}{0.927} \\
\multicolumn{1}{|c|}{\multirow{-9}{*}{\textbf{\begin{tabular}[c]{@{}c@{}}Fake News \\ Detection\end{tabular}}}} & \multicolumn{1}{c|}{\multirow{-3}{*}{\textbf{ANTiVax}}} & F1 & \multicolumn{1}{c|}{0.872} & \multicolumn{1}{c|}{0.865} & 0.908 & \multicolumn{1}{c|}{0.833} & \multicolumn{1}{c|}{0.935} & \multicolumn{1}{c|}{\underline{ 0.942}} & 0.939 & \multicolumn{1}{c|}{0.937} & \multicolumn{1}{c|}{\textbf{0.945}} \\ \hline
\multicolumn{1}{|c|}{} & \multicolumn{1}{c|}{} & Pre. & \multicolumn{1}{c|}{0.531} & \multicolumn{1}{c|}{0.652} & \underline{ 0.686} & \multicolumn{1}{c|}{0.658} & \multicolumn{1}{c|}{0.667} & \multicolumn{1}{c|}{0.646} & 0.647 & \multicolumn{1}{c|}{0.712} & \multicolumn{1}{c|}{\textbf{0.748}} \\
\multicolumn{1}{|c|}{} & \multicolumn{1}{c|}{} & Rec. & \multicolumn{1}{c|}{0.529} & \multicolumn{1}{c|}{0.697} & \underline{ 0.699} & \multicolumn{1}{c|}{0.681} & \multicolumn{1}{c|}{0.672} & \multicolumn{1}{c|}{0.651} & 0.661 & \multicolumn{1}{c|}{0.703} & \multicolumn{1}{c|}{\textbf{0.718}} \\
\multicolumn{1}{|c|}{\multirow{-3}{*}{\textbf{\begin{tabular}[c]{@{}c@{}}Hate Speech \\ Detection\end{tabular}}}} & \multicolumn{1}{c|}{\multirow{-3}{*}{\textbf{HASOC}}} & F1 & \multicolumn{1}{c|}{0.591} & \multicolumn{1}{c|}{0.634} & \underline{ 0.698} & \multicolumn{1}{c|}{0.614} & \multicolumn{1}{c|}{0.657} & \multicolumn{1}{c|}{0.641} & 0.648 & \multicolumn{1}{c|}{0.702} & \multicolumn{1}{c|}{\textbf{0.713}} \\ \hline
\multicolumn{1}{|c|}{} & \multicolumn{1}{c|}{} & Pre. & \multicolumn{1}{c|}{0.785} & \multicolumn{1}{c|}{0.816} & 0.846 & \multicolumn{1}{c|}{0.821} & \multicolumn{1}{c|}{0.841} & \multicolumn{1}{c|}{\underline{ 0.861}} & 0.852 & \multicolumn{1}{c|}{0.854} & \multicolumn{1}{c|}{\textbf{0.877}} \\
\multicolumn{1}{|c|}{} & \multicolumn{1}{c|}{} & Rec. & \multicolumn{1}{c|}{0.783} & \multicolumn{1}{c|}{0.791} & 0.841 & \multicolumn{1}{c|}{0.779} & \multicolumn{1}{c|}{0.842} & \multicolumn{1}{c|}{\underline{ 0.862}} & 0.851 & \multicolumn{1}{c|}{0.843} & \multicolumn{1}{c|}{\textbf{0.875}} \\
\multicolumn{1}{|c|}{} & \multicolumn{1}{c|}{\multirow{-3}{*}{\textbf{Pheme}}} & F1 & \multicolumn{1}{c|}{0.782} & \multicolumn{1}{c|}{0.801} & 0.844 & \multicolumn{1}{c|}{0.799} & \multicolumn{1}{c|}{0.841} & \multicolumn{1}{c|}{\underline{ 0.861}} & 0.852 & \multicolumn{1}{c|}{0.844} & \multicolumn{1}{c|}{\textbf{0.875}} \\ \cline{2-12} 
\multicolumn{1}{|c|}{} & \multicolumn{1}{c|}{} & Pre. & \multicolumn{1}{c|}{0.866} & \multicolumn{1}{c|}{0.824} & 0.928 & \multicolumn{1}{c|}{\underline{ 0.929}} & \multicolumn{1}{c|}{0.851} & \multicolumn{1}{c|}{0.899} & 0.913 & \multicolumn{1}{c|}{0.943} & \multicolumn{1}{c|}{\textbf{0.961}} \\
\multicolumn{1}{|c|}{} & \multicolumn{1}{c|}{} & Rec. & \multicolumn{1}{c|}{0.794} & \multicolumn{1}{c|}{0.829} & \underline{ 0.954} & \multicolumn{1}{c|}{0.839} & \multicolumn{1}{c|}{0.849} & \multicolumn{1}{c|}{0.891} & 0.909 & \multicolumn{1}{c|}{0.937} & \multicolumn{1}{c|}{\textbf{0.968}} \\
\multicolumn{1}{|c|}{} & \multicolumn{1}{c|}{\multirow{-3}{*}{\textbf{Twitter15}}} & F1 & \multicolumn{1}{c|}{0.811} & \multicolumn{1}{c|}{0.825} & \underline{ 0.941} & \multicolumn{1}{c|}{0.881} & \multicolumn{1}{c|}{0.848} & \multicolumn{1}{c|}{0.891} & 0.908 & \multicolumn{1}{c|}{0.936} & \multicolumn{1}{c|}{\textbf{0.957}} \\ \cline{2-12} 
\multicolumn{1}{|c|}{} & \multicolumn{1}{c|}{} & Pre. & \multicolumn{1}{c|}{0.871} & \multicolumn{1}{c|}{0.759} & 0.901 & \multicolumn{1}{c|}{\underline{ 0.941}} & \multicolumn{1}{c|}{0.892} & \multicolumn{1}{c|}{0.921} & 0.895 & \multicolumn{1}{c|}{0.944} & \multicolumn{1}{c|}{\textbf{0.946}} \\
\multicolumn{1}{|c|}{} & \multicolumn{1}{c|}{} & Rec. & \multicolumn{1}{c|}{0.751} & \multicolumn{1}{c|}{0.763} & \textbf{0.942} & \multicolumn{1}{c|}{0.842} & \multicolumn{1}{c|}{0.888} & \multicolumn{1}{c|}{0.918} & 0.891 & \multicolumn{1}{c|}{0.936} & \multicolumn{1}{c|}{\underline{0.937}} \\
\multicolumn{1}{|c|}{} & \multicolumn{1}{c|}{\multirow{-3}{*}{\textbf{Twitter16}}} & F1 & \multicolumn{1}{c|}{0.778} & \multicolumn{1}{c|}{0.759} & 0.919 & \multicolumn{1}{c|}{0.889} & \multicolumn{1}{c|}{0.887} & \multicolumn{1}{c|}{0.919} & 0.892 & \multicolumn{1}{c|}{0.937} & \multicolumn{1}{c|}{\textbf{0.938}} \\ \cline{2-12} 
\multicolumn{1}{|c|}{} & \multicolumn{1}{c|}{} & Pre. & \multicolumn{1}{c|}{0.545} & \multicolumn{1}{c|}{0.583} & 0.571 & \multicolumn{1}{c|}{0.655} & \multicolumn{1}{c|}{0.631} & \multicolumn{1}{c|}{0.556} & 0.602 & \multicolumn{1}{c|}{\textbf{0.746}} & \multicolumn{1}{c|}{\underline{0.721}} \\
\multicolumn{1}{|c|}{} & \multicolumn{1}{c|}{} & Rec. & \multicolumn{1}{c|}{0.676} & \multicolumn{1}{c|}{\underline{0.777}} & \textbf{0.888} & \multicolumn{1}{c|}{0.444} & \multicolumn{1}{c|}{0.555} & \multicolumn{1}{c|}{0.533} & 0.602 & \multicolumn{1}{c|}{0.686} & \multicolumn{1}{c|}{0.718} \\
\multicolumn{1}{|c|}{\multirow{-12}{*}{\textbf{\begin{tabular}[c]{@{}c@{}}Rumour \\ Detection\end{tabular}}}} & \multicolumn{1}{c|}{\multirow{-3}{*}{\textbf{\begin{tabular}[c]{@{}c@{}}Rumour\\ Eval\end{tabular}}}} & F1 & \multicolumn{1}{c|}{0.598} & \multicolumn{1}{c|}{0.667} & \underline{ 0.695} & \multicolumn{1}{c|}{0.518} & \multicolumn{1}{c|}{0.573} & \multicolumn{1}{c|}{0.533} & 0.595 & \multicolumn{1}{c|}{0.697} & \multicolumn{1}{c|}{\textbf{0.712}} \\ \hline
\multicolumn{1}{|c|}{} & \multicolumn{1}{c|}{} & Pre. & \multicolumn{1}{c|}{0.701} & \multicolumn{1}{c|}{0.741} & 0.751 & \multicolumn{1}{c|}{0.734} & \multicolumn{1}{c|}{0.758} & \multicolumn{1}{c|}{0.797} & \underline{ 0.822} & \multicolumn{1}{c|}{0.811} & \multicolumn{1}{c|}{\textbf{0.823}} \\
\multicolumn{1}{|c|}{} & \multicolumn{1}{c|}{} & Rec. & \multicolumn{1}{c|}{0.669} & \multicolumn{1}{c|}{0.746} & 0.751 & \multicolumn{1}{c|}{0.718} & \multicolumn{1}{c|}{0.742} & \multicolumn{1}{c|}{\underline{ 0.798}} & 0.796 & \multicolumn{1}{c|}{0.802} & \multicolumn{1}{c|}{\textbf{0.832}} \\
\multicolumn{1}{|c|}{} & \multicolumn{1}{c|}{\multirow{-3}{*}{\textbf{\begin{tabular}[c]{@{}c@{}}FigLang \\ Twitter\end{tabular}}}} & F1 & \multicolumn{1}{c|}{0.681} & \multicolumn{1}{c|}{0.741} & 0.752 & \multicolumn{1}{c|}{0.721} & \multicolumn{1}{c|}{0.757} & \multicolumn{1}{c|}{0.797} & \underline{ 0.801} & \multicolumn{1}{c|}{0.812} & \multicolumn{1}{c|}{\textbf{0.822}} \\ \cline{2-12} 
\multicolumn{1}{|c|}{} & \multicolumn{1}{c|}{} & Pre. & \multicolumn{1}{c|}{0.595} & \multicolumn{1}{c|}{0.672} & 0.679 & \multicolumn{1}{c|}{0.671} & \multicolumn{1}{c|}{0.639} & \multicolumn{1}{c|}{\textbf{0.723}} & 0.691 & \multicolumn{1}{c|}{0.707} & \multicolumn{1}{c|}{\underline{0.712}} \\
\multicolumn{1}{|c|}{} & \multicolumn{1}{c|}{} & Rec. & \multicolumn{1}{c|}{0.605} & \multicolumn{1}{c|}{0.677} & 0.683 & \multicolumn{1}{c|}{0.664} & \multicolumn{1}{c|}{0.634} & \multicolumn{1}{c|}{\underline{ 0.696}} & 0.688 & \multicolumn{1}{c|}{0.697} & \multicolumn{1}{c|}{\textbf{0.704}} \\
\multicolumn{1}{|c|}{\multirow{-6}{*}{\textbf{\begin{tabular}[c]{@{}c@{}}Sarcasm \\ Detection\end{tabular}}}} & \multicolumn{1}{c|}{\multirow{-3}{*}{\textbf{\begin{tabular}[c]{@{}c@{}}FigLang \\ Reddit\end{tabular}}}} & F1 & \multicolumn{1}{c|}{0.585} & \multicolumn{1}{c|}{0.667} & 0.678 & \multicolumn{1}{c|}{0.665} & \multicolumn{1}{c|}{0.631} & \multicolumn{1}{c|}{0.677} & \underline{ 0.689} & \multicolumn{1}{c|}{ 0.698} & \multicolumn{1}{c|}{ \textbf{0.701}} \\ \hline
\end{tabular}}

\caption{Performance comparisons (Precision (Pre.), Recall (Rec.) and F1 score) of various baselines against $\texttt{\modelName-}\texttt{hyperbolic}$ (Hyper.) and $\texttt{\modelName-}\texttt{euclidean}$ (Eucli.). The best  ({\em resp.} 2nd ranked) method is marked in bold ({\em resp.} underline). See Table \ref{tab:data_stats} for other abbreviations.}
 \vspace{-5mm}
\end{table}

\textbf{Curvature for our implementation}.  \modelName\ learns the hyperbolic representations for public discourse and source-post text simulataneously and applies a novel Fourier co-attention mechanism over the obtained embeddings. However, to be able to do so, we need to ensure that the curvatures of the hyperbolic manifolds (in our case \textit{Poincar\'e ball} model) are same (or a product space of both manifolds). To ensure consistency across both the pipelines (public discourse encoder (Section \ref{sec:comment_encoder}) and source-post encoder (Section \ref{sec:content_encoder})), in \modelName\ we take the constant negative curvature $c = -1$. As addressed in the limitations (See Section \ref{sec:conclusion}), another promising approach for \modelName\ could be to consider the product space of both the manifolds before applying the co-attention mechanism. 

\textbf{Baseline methods.}\
We compare \modelName\ with two sets of baselines (c.f. Table \ref{tab:data_stats}) -- {\bf (i)  Generic neural baselines:} We employ those models  that are often used for social-text classification tasks and have been shown to perform comparatively. We consider different variations of the Transformer model and those who use social context as an auxiliary signal for social-text classification (dEFEND \cite{shu2019defend}). \textbf{ (ii) Data-specific baselines}: We experimented with many data-specific and task-specific baselines and chose top three for every dataset based on the performance. 
Since top three models are data-specific, we call them with generic names -- (a) SOTA-1, (b) SOTA-2, and (c) SOTA-3, respectively.

\textbf{Performance comparison.}
Table \ref{tab:comparison} shows the performance comparison.
The content-based pre-trained models, {BERT} and {RoBERTa}, outperform {dEFEND} which uses both the source content and user comments.
We observe that {dEFEND} performs better than all the data-specific baselines because of the sophisticated use of co-attention. By incorporating public comments along with the social post, \modelName\ shows significant\footnote{We also perform statistical significance $t$-test comparing \modelName\ and the other baselines.} performance improvement over all the baselines. 
We observe that while the performance improvement over baselines is significant ($\sim4\%$; $p<0.005$) on datasets like  Politifact, Gossipcop, and Twitter15,    the performance improvement  is not that significant ($p<0.05$) on AntiVax and FigLang (Reddit). This is due to the fact that in the latter datasets, there are less number of comments available per the source posts (see Table \ref{tab:data_stats}). On Politifact and Gossipcop, \modelName-\texttt{hyperbolic} has a performance gain of $3.9\%$ and $3.8\%$, over the best baselines models, {RoBERTa} and {dEFEND}, respectively. Note that even when compared to the pre-trained Transformer architectures, \texttt{Hpyhen} shows decent improvement, while for the non-Transformer based baselines like \texttt{HAN}, there is a performance gain of $11.2\%$ even on the AntiVax dataset. We explain the data-specific baselines, their modalities, and detailed analyses of their performance in the {\color{black}{Appendix \ref{appendix:experiments}}}.

\textbf{Ablation study.} 
We perform ablations with two variants of our model, namely 
$\texttt{\modelName-}\texttt{hyperbolic}$ and $\texttt{\modelName-}\texttt{euclidean}$, in which  {Hyperbolic} and  {Euclidean} represent the underlying manifold.\\
{\bf $\blacksquare$ Effect of public wisdom.} When we remove user comments (\modelName\ {\tt w/o comments}: 
we consider only source post, get rid of the co-attention block for this analysis as we have just one modality, and keep the Fourier transform layer to capture the latent messages in the candidate post), the performance degrades. Table \ref{tab:ablation} shows that for {Gossipcop} and {Politifact} datasets, \modelName-\texttt{hyperbolic} has a performance degradation of $7.23\%$ and $7.4\%$, respectively. Due to the presence of less number of comments per post in the {AntiVax} dataset, the performance degradation is not that significant (i.e., $1.05\%$, $p<0.1$). On some datasets like FigLang (Twitter), Pheme and Twitter15, \modelName-\texttt{hyperbolic w/o comments} records a significant performance degradation ($p < 0.001$) of $8.47\%$, $7.4\%$ and $6.24\%$, respectively. Even \modelName-\texttt{euclidean w/o comments} sees a fall in F1 score of $6.38\%$, $6.4\%$ and $5.45\%$ for Twitter15, Twitter16 and FigLang (Twitter), respectively. Since this is a content-only pipeline, in many cases, the model is outperformed by pre-trained Transformer models.\\
$\blacksquare$ \textbf{Effect of hyperbolic space.} 
We evaluate \modelName's performance by replacing the hyperbolic manifold with Euclidean. 
We observe that in support of our initial hypothesis, \modelName-\texttt{hyperbolic} outperforms \modelName-\texttt{euclidean} (see Table \ref{tab:ablation}). The former records a considerable gain of $3.55\%$ and $2.85\%$ F1 score on Gossipcop and Politifact datasets, respectively, over the latter. For the AntiVax dataset, a smaller increment of $0.83\%$ can be attributed to the less number of user comments available in the dataset. Note that for the variant, \modelName-\texttt{hyperbolic} {\tt w/o comments}, there is a performance degradation as compared to \modelName-\texttt{euclidean} on Gossipcop and Politifact. This is intuitive as the sole advantage of hyperbolic space lies in capturing the inherent hierarchy of the macro-AMR graphs. Therefore, in case of a content-only model, \modelName-\texttt{euclidean} performs better. On Pheme and Twitter15, the former achieves a significant F1 score gain ($p < 0.005$) of $3.12\%$ and $3.18\%$ respectively. Due to less number of user comments in RumourEval, FigLang (Twitter) and Figlang (Reddit), the performance gain is less significant ($p < 0.05$), i.e., $1.44\%$, $1.01\%$ and $0.47\%$ respectively. It should be noted that this behaviour demonstrates the effectiveness of \modelName\ in \textit{early detection}. Even with less number of user comments available, \modelName\ achieves performance boost over the baselines, and thus can be extremely effective in tasks like detecting fake news, where \textit{early detection} is of great significance.\\ 
$\blacksquare$ \textbf{Effect of Fourier transform layer.}  Table \ref{tab:ablation} shows that including the Fourier transform layer to capture the most prominent user opinions about the source post and the most common (latent) messages conveyed by the source post, boosts the overall performance of \modelName. There is an improvement of $5.6\%$  F1 score on Gossipcop and $1.74\%$ on Politifact due to the Fourier layer in \modelName-\texttt{hyperbolic}. Because of the less number of comments per post in AntiVax, there is a smaller increment of $0.87\%$  F1 score. Even for \modelName-\texttt{euclidean}, there is an increase of $2.28\%$ on Gossipcop and $\sim1$\% on Politifact. \modelName-\texttt{hyperbolic} shows a significant improvement ($p < 0.001$) of $4.49\%$, $4.34\%$, and $4.13\%$ F1 score on FigLang (Twitter), Pheme and Twitter15, respectively. For \modelName-\texttt{euclidean}, there is an increase of $5.10\%$ on FigLang (Twitter) and $3.53\%$ on FigLang (Reddit). \modelName-\texttt{euclidean} and \modelName-\texttt{hyperbolic} record an average increase of $4.49\%$ and $4.34\%$ in F1 score, respectively, over all datasets. On applying co-attention over the outputs of Fourier transform layer, we are able to attend better to both the representations simultaneously, and thus the model's ability to capture the correlation between the two increases.

\begin{table}[]
\resizebox{\columnwidth}{!}{%
\begin{tabular}{|
>{\columncolor[HTML]{FFFFFF}}c |
>{\columncolor[HTML]{FFFFFF}}l |
>{\columncolor[HTML]{FFFFFF}}c 
>{\columncolor[HTML]{FFFFFF}}c 
>{\columncolor[HTML]{FFFFFF}}c |
>{\columncolor[HTML]{FFFFFF}}c 
>{\columncolor[HTML]{FFFFFF}}c 
>{\columncolor[HTML]{FFFFFF}}c |}
\hline
\cellcolor[HTML]{FFFFFF} & \multicolumn{1}{c|}{\cellcolor[HTML]{FFFFFF}} & \multicolumn{3}{c|}{\cellcolor[HTML]{FFFFFF}\textbf{Euclidean}} & \multicolumn{3}{c|}{\cellcolor[HTML]{FFFFFF}\textbf{Hyperbolic}} \\ \cline{3-8} 
\multirow{-2}{*}{\cellcolor[HTML]{FFFFFF}\textbf{Dataset}} & \multicolumn{1}{c|}{\multirow{-2}{*}{\cellcolor[HTML]{FFFFFF}\textbf{Model}}} & \multicolumn{1}{c|}{\cellcolor[HTML]{FFFFFF}\textbf{Precision}} & \multicolumn{1}{c|}{\cellcolor[HTML]{FFFFFF}\textbf{Recall}} & \textbf{F1} & \multicolumn{1}{c|}{\cellcolor[HTML]{FFFFFF}\textbf{Precision}} & \multicolumn{1}{c|}{\cellcolor[HTML]{FFFFFF}\textbf{Recall}} & \textbf{F1} \\ \hline
\cellcolor[HTML]{FFFFFF} & \modelName\ & \textbf{0.9515} & \textbf{0.9364} & \textbf{0.9401} & \textbf{0.9722} & \textbf{0.9612} & \textbf{0.9686} \\
\cellcolor[HTML]{FFFFFF} & \modelName\ \texttt{w/o comments} & 0.9166 & 0.8802 & 0.8979 ($\downarrow$ 4.22\%) & 0.8461 & 0.9615 & 0.8963 ($\downarrow$ 7.23\%) \\
\multirow{-3}{*}{\cellcolor[HTML]{FFFFFF}\textbf{Politifact}} & \modelName\ \texttt{w/o Fourier} & 0.9091 & 0.9523 & 0.9302 ($\downarrow$ 0.99\%) & 0.9341 & 0.9623 & 0.9512 ($\downarrow$ 1.74\%) \\ \hline
\cellcolor[HTML]{FFFFFF} & \modelName\ & \textbf{0.7862} & \textbf{0.7763} & \textbf{0.7812} & \textbf{0.7913} & \textbf{0.7884} & \textbf{0.8167} \\
\cellcolor[HTML]{FFFFFF} & \modelName\ \texttt{w/o comments} & 0.7557 & 0.7578 & 0.7551 ($\downarrow$ 2.61\%) & 0.7511 & 0.7734 & 0.7407 ($\downarrow$ 7.60\%) \\
\multirow{-3}{*}{\cellcolor[HTML]{FFFFFF}\textbf{Gossipcop}} & \modelName\ \texttt{w/o Fourier} & 0.7751 & 0.7695 & 0.7584 ($\downarrow$ 2.28\%) & 0.7611 & 0.7812 & 0.7607 ($\downarrow$ 5.60\%) \\ \hline
\cellcolor[HTML]{FFFFFF} & \modelName\ & \textbf{0.9409} & \textbf{0.9375} & \textbf{0.9373} & \textbf{0.9511} & \textbf{0.9275} & \textbf{0.9456} \\
\cellcolor[HTML]{FFFFFF} & \modelName\ \texttt{w/o comments} & 0.9202 & 0.9187 & 0.9192 ($\downarrow$ 1.81\%) & 0.9417 & 0.9346 & 0.9351 ($\downarrow$ 1.05\%) \\
\multirow{-3}{*}{\cellcolor[HTML]{FFFFFF}\textbf{ANTiVax}} & \modelName\ \texttt{w/o Fourier} & 0.9315 & 0.9281 & 0.9286 ($\downarrow$ 0.87\%) & 0.9365 & 0.9281 & 0.9369 ($\downarrow$ 0.87\%) \\ \hline
\cellcolor[HTML]{FFFFFF} & \modelName\ & \textbf{0.7121} & \textbf{0.7031} & \textbf{0.7031} & \textbf{0.7481} & \textbf{0.7187} & \textbf{0.7132} \\
\cellcolor[HTML]{FFFFFF} & \modelName\ \texttt{w/o comments} & 0.7122 & 0.6718 & 0.6693 ($\downarrow$ 3.38\%) & 0.6747 & 0.6718 & 0.6717 ($\downarrow$ 4.15\%) \\
\multirow{-3}{*}{\cellcolor[HTML]{FFFFFF}\textbf{HASOC}} & \modelName\ \texttt{w/o Fourier} & 0.6909 & 0.6718 & 0.6762 ($\downarrow$ 2.69\%) & 0.7019 & 0.7031 & 0.6933 ($\downarrow$ 1.99\%) \\ \hline
\cellcolor[HTML]{FFFFFF} & \modelName\ & \textbf{0.8545} & \textbf{0.8437} & \textbf{0.8445} & \textbf{0.8771} & \textbf{0.8751} & \textbf{0.8757} \\
\cellcolor[HTML]{FFFFFF} & \modelName\ \texttt{w/o comments} & 0.8264 & 0.8142 & 0.8161 ($\downarrow$ 2.84\%) & 0.8121 & 0.7968 & 0.8017 ($\downarrow$ 7.40\%) \\
\multirow{-3}{*}{\cellcolor[HTML]{FFFFFF}\textbf{Pheme}} & \modelName\ \texttt{w/o Fourier} & 0.8304 & 0.8203 & 0.8215 ($\downarrow$ 2.30\%) & 0.8411 & 0.8301 & 0.8323 ($\downarrow$ 4.34\%) \\ \hline
\cellcolor[HTML]{FFFFFF} & \modelName\ & \textbf{0.9437} & \textbf{0.9375} & \textbf{0.9367} & \textbf{0.9703} & \textbf{0.9687} & \textbf{0.9685} \\
\cellcolor[HTML]{FFFFFF} & \modelName\ \texttt{w/o comments} & 0.8782 & 0.8751 & 0.8729 ($\downarrow$ 6.38\%) & 0.9078 & 0.9062 & 0.9061 ($\downarrow$ 6.24\%) \\
\multirow{-3}{*}{\cellcolor[HTML]{FFFFFF}\textbf{Twitter15}} & \modelName\ \texttt{w/o Fourier} & 0.9082 & 0.9062 & 0.9063 ($\downarrow$ 3.04\%) & 0.9444 & 0.9375 & 0.9272 ($\downarrow$ 4.13\%) \\ \hline
\cellcolor[HTML]{FFFFFF} & \modelName\ & \textbf{0.9444} & \textbf{0.9363} & \textbf{0.9372} & \textbf{0.9464} & \textbf{0.9375} & \textbf{0.9382} \\
\cellcolor[HTML]{FFFFFF} & \modelName\ \texttt{w/o comments} & 0.9021 & 0.8751 & 0.8732 ($\downarrow$ 6.40\%) & 0.9196 & 0.9061 & 0.9042 ($\downarrow$ 3.40\%) \\
\multirow{-3}{*}{\cellcolor[HTML]{FFFFFF}\textbf{Twitter16}} & \modelName\ \texttt{w/o Fourier} & 0.9211 & 0.9071 & 0.9054 ($\downarrow$ 3.18\%) & 0.9067 & 0.9062 & 0.9155 ($\downarrow$ 2.27\%) \\ \hline
\cellcolor[HTML]{FFFFFF} & \modelName\ & \textbf{0.7465} & \textbf{0.6862} & \textbf{0.6979} & \textbf{0.7219} & \textbf{0.7187} & \textbf{0.7123} \\
\cellcolor[HTML]{FFFFFF} & \modelName\ \texttt{w/o comments} & 0.6776 & 0.6364 & 0.6611 ($\downarrow$ 3.68\%) & 0.6941 & 0.6875 & 0.6898 ($\downarrow$ 2.25\%) \\
\multirow{-3}{*}{\cellcolor[HTML]{FFFFFF}\textbf{RumourEval}} & \modelName\ \texttt{w/o Fourier} & 0.7045 & 0.6875 & 0.6762 ($\downarrow$ 2.17\%) & 0.7433 & 0.6875 & 0.6743 ($\downarrow$ 3.80\%) \\ \hline
\cellcolor[HTML]{FFFFFF} & \modelName\ & \textbf{0.8115} & \textbf{0.8025} & \textbf{0.8121} & \textbf{0.8235} & \textbf{0.8321} & \textbf{0.8222} \\
\cellcolor[HTML]{FFFFFF} & \modelName\ \texttt{w/o comments} & 0.7656 & 0.7583 & 0.7576 ($\downarrow$ 5.45\%) & 0.7555 & 0.7375 & 0.7375 ($\downarrow$ 8.47\%) \\
\multirow{-3}{*}{\cellcolor[HTML]{FFFFFF}\textbf{FigLang\_Twitter}} & \modelName\ \texttt{w/o Fourier} & 0.7624 & 0.7617 & 0.7611 ($\downarrow$ 5.10\%) & 0.7779 & 0.7968 & 0.7773 ($\downarrow$ 4.49\%) \\ \hline
\cellcolor[HTML]{FFFFFF} & \modelName\ & \textbf{0.7071} & \textbf{0.6979} & \textbf{0.6971} & \textbf{0.7107} & \textbf{0.7043} & \textbf{0.7018} \\
\cellcolor[HTML]{FFFFFF} & \modelName\ \texttt{w/o comments} & 0.6685 & 0.6511 & 0.6489 ($\downarrow$ 4.82\%) & 0.6743 & 0.6514 & 0.6513 ($\downarrow$ 5.05\%) \\
\multirow{-3}{*}{\cellcolor[HTML]{FFFFFF}\textbf{FigLang\_Reddit}} & \modelName\ \texttt{w/o Fourier} & 0.6687 & 0.6642 & 0.6618 ($\downarrow$ 3.53\%) & 0.7091 & 0.6971 & 0.6961 ($\downarrow$ 0.57\%) \\ \hline
\end{tabular}%
}
\caption{Ablation study showing the effect of public discourse, hyperbolic manifold, and Fourier transform layer on the performance of \modelName\ for all four tasks. The decrease in performance of the ablation version of \modelName\ w.r.t its original one is shown within parenthesis.}
\label{tab:ablation}
\end{table}


%% file: sections/conclusion.tex
\vspace{-3mm}
\section{Explainability}\label{sec:exp}
Here, we demonstrate how \modelName\ excels at providing explanations for social-text classification tasks. Using the hyperbolic co-attention weights $\mathbf{a}^{s\mathcal{E}}$ (Equation \ref{eq:attention_weights}), we can provide an implicit rank list of sentences present in the source post in the order of their relevance to the final prediction. For instance, consider the scenario of fake news detection. A fake news is often created by manipulating selected parts of a true information. The generated rank list of sentences in this case would correspond to the sentences in the news article, which are possible misinformation. Furthermore, manual verification of all sentences in a news article is tedious, and therefore, a rank list based on the level of check-worthiness of sentences is convenient. 
\begin{wraptable}{r}{6.8cm}
\centering
\scriptsize
\begin{tabular}{|l|l|l|}
\hline
\textbf{Model} & \textbf{\textit{Kendall's}} $\tau$& \textbf{\textit{Spearman's}} $\rho$\\ \hline
dEFEND & 0.0231 $\pm$\ 0.053 & 0.0189 $\pm$\ 0.012 \\ 
\modelName-\texttt{euclidean} & 0.4013 $\pm$\ 0.072 & 0.4236 $\pm$\ 0.072 \\
\modelName-\texttt{hyperbolic} & \textbf{0.4983 $\pm$\ 0.055} & \textbf{0.5532} $\pm$\ \textbf{0.045} \\ \hline
\end{tabular}
\caption{Performance of \modelName\ and dEFEND in providing explanations on Politifact. $\pm$\ denotes std. dev. across  5  random runs.}
\label{tab:exp}
\end{wraptable}
To evaluate the performance of \modelName\ in generating explanations, we consider Politifact, and for each source post, we manually annotate sentences present in the source post based on their relevance to the final level (fake/real) (see \textcolor{black}{Appendix} \ref{appendix:explainability} for dataset annotation details). The annotators also rank sentences of a source post in the order of their check-worthiness. We expect our model to produce a similar list of sentences for the source post.
We use dEFEND \cite{shu2019defend} as a baseline for comparing the rank correlations, and evaluate the rank list produced by \modelName\ against this ground-truth \textit{annotated rank list} using Kendall's $\tau$ and Spearman's $\rho$ rank correlation coefficients. dEFEND is the only model among the chosen baselines, which produces a similar rank link using attention weights in an attempt to provide explanations (See {\color{black}Appendix \ref{appendix:explainability}} for sample rank lists generated by dEFEND and \modelName). Table \ref{tab:exp} shows that the explanations produced by dEFEND have almost no correlation $(\tau = 0.0231, \rho = 0.0189)$ to the annotated rank list. On the contrary, \modelName-\texttt{hyperbolic} shows a high positive correlation $(\tau = 0.4983, \rho = 0.5532)$. \modelName-\texttt{euclidean} also shows comparable performance. 
The results present the efficacy of \modelName\ in providing decent explanations for social-text classification.

\textbf{Model augmentation for explainability.} To provide explanations, we rule out the Fourier sub-layer from \modelName. This is done because on taking the Fourier transform of source-post and comments' representations (Equation \ref{eq:fourier_layer}), we cannot assert an ordered mapping from the spectral domain to the sentence representations. Such an order is necessary for us to have a mapping between the co-attention weights and the source-post sentences they were derived from. Without such a mapping, \modelName\ would not be able to generate a rank-list based on the sentences in the source-post.

\vspace{-3.1mm}
\section{Conclusion}
\label{sec:conclusion}
Public wisdom on social media  carries diverse latent signals which can be used in unison with the source post to enhance the social-text classification tasks. Our proposed \modelName\ model uses a novel hyperbolic Fourier co-attention network to amalgamate both these information. Apart from the state-of-the-art performance in social-text classification, \modelName\ shows the potential of generating suitable explanations to support the final prediction and works well in a \textit{generalised} discourse-aware setting. In the future, mixed-curvature learning in product spaces \cite{gu2018learning} and hyperbolic-to-hyperbolic \cite{dai2021hyperbolic} representations could be employed to boost the learning capabilities of \modelName.

\textbf{Limitations}. \modelName\ resorts to using tangent spaces for computing Fourier co-attention which is inferior because tangent spaces are only a local approximation of the manifold. One may further incorporate other signals such as user interaction network and user credibility into the model. 

\section*{Acknowledgment}
T. Chakraborty would like to acknowledge the support of the LinkedIn faculty research grant. We would like to acknowledge the support of the data annotators - Arnav Goel, Samridh Girdhar, Abhijay Singh, and Siddharth Rajput, for their help in annotating the Politifact dataset.

%% file: sections/appendix.tex
\newpage
\appendix

\section{Background} 

\subsection{Hyperbolic Space Models}\label{hyper-models}
Hyperbolic space has been studied under five isometric models \cite{cannon1997hyperbolic}. In this section, we discuss the \textit{Poincar\'e ball} and \textit{Lorentz} models, which are utilized in the source-post encoder. \modelName\ relies on the \textit{Poincar\'e ball} model.

\textbf{Poincar\'e ball model}. 
The Poincar\'e ball $\left(\mathbb{B}^d_c, g^{\mathbb{B}}\right)$ of radius $1/\sqrt{|c|}$, equipped with Riemannian metric $g^{\mathcal{B}}$ and constant negative curvature $c\ (c<0)$, is a $d$-dimensional manifold ${\mathbb{B}_c^d = \{\mathbf{x} \in \mathbb{R}^d: c\|\mathbf{x}\|^2 < -1}\}$, where $g^{\mathcal{B}}$ is \textit{conformal} to the Euclidean metric $g^{\mathcal{E}} = \mathbf{I}_d$ with \textit{conformal} factor $\lambda_{\mathbf{x}}^c = 2 / (1 + c\|\mathbf{x}\|^2)$. The distance between two points $\mathbf{x}, \mathbf{y} \in \mathbb{B}^d_c$ is measured along a \textit{geodesic} and is given by $d_{\mathcal{B}}^c(\mathbf{x}, \mathbf{y}) = (2/\sqrt{|c|})\tanh^{-1}(\sqrt{|c|}\|-\mathbf{x} \oplus_c \mathbf{y}\|)$.

\textbf{Lorentz model}. 
With constant negative curvature $c\ (c < 0)$, and equipped with Riemannian metric $g^{\mathcal{L}}$, the Lorentz model $\left(\mathbb{L}^d_c, g^{\mathbb{L}}\right)$ is the Riemannian manifold $\mathbb{L}_c^d = \{\mathbf{x} \in \mathbb{R}^{d+1}: \langle\mathbf{x}, \mathbf{x}\rangle_{\mathcal{L}} = 1/c\}$, where $g^{\mathcal{L}} = \text{diag}([-1, 1, ..., 1])_n$ .The distance between two points $\mathbf{x}, \mathbf{y} \in \mathbb{L}^d_c$ is given by $d_{\mathcal{L}}^c(\mathbf{x}, \mathbf{y}) = (1/\sqrt{|c|})\ \text{cosh}^{-1}(c\langle\mathbf{x}, \mathbf{y}\rangle_{\mathcal{L}})$, where $\langle\mathbf{x}, \mathbf{y}\rangle_{\mathcal{L}}$ is the Lorentzian inner product.

\textbf{Klein model}. 
With constant negative curvature $c\ (c < 0)$, the Klein model is also the Riemannian manifold ${\mathbb{K}_c^d = \{\mathbf{x} \in \mathbb{R}^d: c\|\mathbf{x}\|^2 < -1}\}$. The isomorphism between the Klein model and Poincar\'e ball can be defined through a projection on or from the hemisphere model.

\subsection{Hyperbolic Graph Convolutional Network (HGCN)} \label{appendix:hgcn}
Given a graph $\mathcal{G}=(\mathcal{V},E)$ and Euclidean input features $(\mathbf{x}_i^{\mathcal{E}})_{i\in\mathcal{V}}$, HGCN \cite{https://doi.org/10.48550/arxiv.1910.12933} can be interpreted as transforming and aggregating neighbours' embeddings in the tangent space of the center node and projecting the result to a hyperbolic space with a different curvature, at each stacked layer. 
Suppose that $\mathbf{x}_i$ aggregates information from its neighbors $(\mathbf{x}_{j})_{j\in\mathcal{N}(i)}$. Further, we use $\mathcal{H}$ in superscript to denote the hyperbolic manifold. Then, the aggregation operation \textsc{Agg} can be formulated as, $\textsc{Agg}^K(\mathbf{x}^{\mathcal{H}})_i = \mathrm{exp}^{K}_{\mathbf{x}^{\mathcal{H}}_i }\left(\sum_{j \in \mathcal{N}(i)} w_{ij} \mathrm{log}^{K}_{\mathbf{x}^{\mathcal{H}}_i}(\mathbf{x}^{\mathcal{H}}_j)\right)$, where, $w_{ij} = \textsc{Softmax}_{j\in\mathcal{N}(i)}(\textsc{MLP}(\mathrm{log}^K_\mathbf{o}(\mathbf{x}_i^{\mathcal{H}}) || \mathrm{log}^K_\mathbf{o}(\mathbf{x}_j^{\mathcal{H}}))) \label{eq:att}$. More precisely, HGCN applies the Euclidean non-linear activation in $\mathcal{T}_\mathbf{o}\mathbb{H}^{d,K_{\ell-1}}$ and then maps back to $\mathbb{H}^{d,K_\ell}$, as $\sigma^{\otimes^{K_{\ell-1},K_{\ell}}}(\mathbf{x}^{\mathcal{H}}) = \mathrm{exp}^{K_{\ell}}_\mathbf{o}(\sigma(\mathrm{log}^{K_{\ell-1}}_\mathbf{o}(\mathbf{x}^{\mathcal{H}}))).$ Therefore, the message passing in a HGCN layer can be shown as, $
    \mathbf{x}^{\ell,\mathcal{H}}_i =  \sigma^{\otimes_{K_{\ell-1},K_\ell}}(\textsc{Agg}^{K_{\ell-1}}((W^{\ell}\otimes_{K_{\ell-1}}\mathbf{x}_i^{\ell-1,\mathcal{H}})\oplus_{K_{\ell-1}}\mathbf{b}^\ell))$,
where $-1/K_{\ell-1}$ and $-1/K_{\ell}$ are the hyperbolic curvatures at layer $\ell-1$ and $\ell$, respectively. As a final step, we can use the hyperbolic node embeddings at the last layer $(\mathbf{x}_i^{L,\mathcal{H}})_{i\in\mathcal{V}}$ for downstream tasks.
\subsection{Hyperbolic Hierarchical Attention Network (HyperHAN)}\label{appendix:hhan}
HyperHAN \cite{zhang2021hype} learns the source document representation through a hierarchical attention network in the hyperbolic space. Consider the input embedding of the $t^{th}$ word appearing in the $i^{th}$ sentence
as $\mathbf{x}_{it}$, in the candidate document. The Euclidean hidden state of $\mathbf{x}_{it}$ within the sentence can be constructed using forward and backward Euclidean-GRU layers as: $\mathbf{h}^{\mathcal{E}}_{\textit{it}} = [\overrightarrow{GRU}(\mathbf{x}_{it}),\overleftarrow{GRU}(\mathbf{x}_{it})]$.  We denote the Klein and Lorentz models using $\mathcal{K}$ and $\mathcal{L}$ in superscript, respectively. Zhang and Gao \cite{zhang2021hype} aim to jointly learn a hyperbolic word centroid $\mathbf{c}^\mathcal{L}_{\textit{w}}$ from all the training documents. 
$\mathbf{c}^\mathcal{L}_{\textit{w}}$ can be considered as a baseline for measuring the importance of hyperbolic words based on their mutual distance. To learn $\mathbf{c}^\mathcal{L}_{\textit{w}}$, they consider another layer upon hidden state $\mathbf{h}^\mathcal{E}_{\textit{it}}$ as: $\mathbf{h}^{\mathcal{E}^{\rq}}_{\textit{it}} = tanh(\mathbf{W}_w\mathbf{h}^\mathcal{E}_{\textit{it}} + \mathbf{b}_w)$. 
The next step is activating  $\mathbf{h}^{\mathcal{E}^{\rq}}_{\textit{it}}$ as $\mathbf{h}^{\mathcal{L}^{\rq}}_{\textit{it}}$. The word-level attention weights are then computed as $\alpha_{it}$ by:
$\alpha_{it} = \exp(-\beta_w d_\mathcal{L}(\mathbf{c}^\mathcal{L}_{\textit{w}}, \mathbf{h}^{\mathcal{L}^{\rq}}_{\textit{it}}) - c_w)$. 
After capturing the hyperbolic attention weights, the semantic meaning of words appearing in the same sentences is aggregated via Einstein midpoint:
$\mathbf{s}^{\mathcal{K}w}_i = \sum_{t}\left[\frac{\alpha_{it}\gamma\left(\mathbf{h}^\mathcal{K}_{\textit{it}}\right)}{\sum_{l}\alpha_{il}\gamma\left(\mathbf{h}^\mathcal{K}_{\textit{it}}\right)}\right]$, 
where, $ \gamma\left(\mathbf{h}^\mathcal{K}_{\textit{it}}\right) = \frac{1}{\sqrt{1 - ||\mathbf{h}^\mathcal{K}_{\textit{it}}||^2}} = \frac{1}{\sqrt{1 - \frac{\sinh^2\left({r_{it}}\right)}{\cosh^2\left({r_{it}}\right)}}}$,  $\gamma\left(\mathbf{h}^{\mathcal{K}}_{\textit{it}}\right)$ 
is the so-called Lorentz factor, and $\mathbf{s}^{\mathcal{K}w}_i$ is the learned representation for the $i${th} sentence. Similar to the word-level encoder, \textit{Mobi\"us}-GRU units are utilized with aggregation using Einstein midpoint to encode each sentence in the source post, yielding the final document level representation. 
 
\section{Experiments}
\label{appendix:experiments}
\textbf{Dataset preparation.} In this section, we list out the dataset collection and augmentation procedure. Since we need both source-post text and public discourse information, we augment all the datasets to yield sufficient comments per social media post. \textbf{Politifact} and \textbf{Gossipcop} \cite{shu2020fakenewsnet} were collected from two fact-verification platforms PolitiFact and GossipCop, and contain news content with two labels (fake or real) and social context information. After scraping the tweets corresponding to the news articles in the datasets, we get 837 and 19266 news articles (Politifact and Gossipcop respectively) which have atleast 1 tweet available. 
Finally, we filter the news articles with atleast 3 comments which gives us datasets with 415 and 2813 news articles (source posts) for Politifact and Gossipcop respectively. \textbf{AntiVax} \cite{hayawi2022anti} is a novel Twitter dataset for COVID-19 vaccine misinformation detection, with more than 15,000 tweets annotated as fake or not. We manually scrape the user comments corresponding to the tweets present in the dataset, which resulted in 3797 tweets (2865 real and 932 fake) with atleast one user comment. \textbf{HASOC} \cite{mandl2019overview} was taken from the HASOC 2019 sub-task B with over 3000 tweets (comments and replies) from 82 conversation threads labelled as hate speech or not. Due to the available labels, we consider the top level comments on the 82 conversation threads as separate tweets and the corresponding replies as the public discourse. This yields a dataset with 712 tweets with public discourse (in contrast to the original 82 conversation threads). \textbf{Pheme} \cite{8118443} is a collection of 6425 Twitter rumours and non-rumours conversation threads related to 9 events and each of the samples is annotated as either True, False or Unverified. \textbf{RumourEval} was introduced in SemEval-2019 Task 7, and has 446 twitter and reddit posts belonging to three categories: \textit{real}, \textit{fake} and \textit{unverified} rumour. \cite{gorrell2019semeval}. \textbf{Twitter15} and \textbf{Twitter16} \cite{ma-etal-2018-rumor} consist of source tweets (1490 and 818 resp.) along with the sequence of re-tweet users. We choose only \textit{true} and \textit{fake} rumour labels as the ground truth. Since there is no discourse available, we scrape the user comments corresponding to the posts and filter the tweets with atleast one comment giving us 543 and 362 source-tweets respectively. \textbf{FigLang (Twitter)} and \textbf{FigLang (Reddit)} \cite{ghosh2020report} are FigLang 2020 shared task datasets with 4400 samples each labelled as either sarcasm or not.

\textbf{Experimentation details.} We adopt a pre-trained AMR parser from the AMRLib\footnote{\url{https://amrlib.readthedocs.io/en/latest/}} library and use the \texttt{parse\_xfm\_bart\_base} model to generate the comment-level AMRs. We resolve co-references on the comment-level AMRs using an off-the-shelf model AMRCoref\footnote{\url{https://github.com/bjascob/amr_coref}}, which yields the various co-reference clusters. Finally, we convert all the merged AMRs to the Deep Graph Library\footnote{\url{https://www.dgl.ai/}} (DGL) format. All node embeddings for AMR are initialised using 100D Glove embeddings\footnote{\url{https://nlp.stanford.edu/projects/glove/}}. Data-specific hyperparameters have been laid out in Table \ref{tab:hyperparams}.

\begin{wraptable}{r}{7cm}
\label{tab:hyperparams}
\tiny
\begin{tabular}{|
>{\columncolor[HTML]{FFFFFF}}c |
>{\columncolor[HTML]{FFFFFF}}c 
>{\columncolor[HTML]{FFFFFF}}c |
>{\columncolor[HTML]{FFFFFF}}c 
>{\columncolor[HTML]{FFFFFF}}c |
>{\columncolor[HTML]{FFFFFF}}c |
>{\columncolor[HTML]{FFFFFF}}c |}
\hline
\cellcolor[HTML]{FFFFFF} & \multicolumn{2}{c|}{\cellcolor[HTML]{FFFFFF}\textbf{Euclidean}} & \multicolumn{2}{c|}{\cellcolor[HTML]{FFFFFF}\textbf{Hyperbolic}} & \cellcolor[HTML]{FFFFFF} & \cellcolor[HTML]{FFFFFF} \\ \cline{2-5}
\multirow{-2}{*}{\cellcolor[HTML]{FFFFFF}\textbf{Dataset}} & \multicolumn{1}{c|}{\cellcolor[HTML]{FFFFFF}\textbf{lr}} & \cellcolor[HTML]{FFFFFF}\textbf{\begin{tabular}[c]{@{}c@{}}Batch\\ size\end{tabular}} & \multicolumn{1}{c|}{\cellcolor[HTML]{FFFFFF}\textbf{lr}} & \textbf{\begin{tabular}[c]{@{}c@{}}Batch\\ size\end{tabular}} & \multirow{-2}{*}{\cellcolor[HTML]{FFFFFF}\textbf{\begin{tabular}[c]{@{}c@{}}Max\\ sents\end{tabular}}} & \multirow{-2}{*}{\cellcolor[HTML]{FFFFFF}\textbf{\begin{tabular}[c]{@{}c@{}}Max\\ coms\end{tabular}}} \\ \hline
Politifact & \multicolumn{1}{c|}{\cellcolor[HTML]{FFFFFF}1e-3} & \cellcolor[HTML]{FFFFFF}16 & \multicolumn{1}{c|}{\cellcolor[HTML]{FFFFFF}1e-2} & 16 & 30 & 10 \\ \hline
Gossipcop & \multicolumn{1}{c|}{\cellcolor[HTML]{FFFFFF}2e-3} & 64 & \multicolumn{1}{c|}{\cellcolor[HTML]{FFFFFF}2e-3} & 64 & 50 & 10 \\ \hline
ANTiVax & \multicolumn{1}{c|}{\cellcolor[HTML]{FFFFFF}1e-4} & \cellcolor[HTML]{FFFFFF}64 & \multicolumn{1}{c|}{\cellcolor[HTML]{FFFFFF}1e-2} & 32 & 2 & 8 \\ \hline
HASOC & \multicolumn{1}{c|}{\cellcolor[HTML]{FFFFFF}1e-4} & 16 & \multicolumn{1}{c|}{\cellcolor[HTML]{FFFFFF}1e-3} & 32 & 2 & 9 \\ \hline
Pheme & \multicolumn{1}{c|}{\cellcolor[HTML]{FFFFFF}1e-3} & 64 & \multicolumn{1}{c|}{\cellcolor[HTML]{FFFFFF}1e-2} & 32 & 2 & 17 \\ \hline
Twitter15 & \multicolumn{1}{c|}{\cellcolor[HTML]{FFFFFF}1e-4} & 32 & \multicolumn{1}{c|}{\cellcolor[HTML]{FFFFFF}1e-2} & 32 & 2 & 8 \\ \hline
Twitter16 & \multicolumn{1}{c|}{\cellcolor[HTML]{FFFFFF}1e-4} & 32 & \multicolumn{1}{c|}{\cellcolor[HTML]{FFFFFF}1e-3} & 32 & 2 & 20 \\ \hline
RumourEval & \multicolumn{1}{c|}{\cellcolor[HTML]{FFFFFF}1e-4} & 16 & \multicolumn{1}{c|}{\cellcolor[HTML]{FFFFFF}1e-2} & 32 & 2 & 3 \\ \hline
\begin{tabular}[c]{@{}c@{}}FigLang\\ Twitter\end{tabular} & \multicolumn{1}{c|}{\cellcolor[HTML]{FFFFFF}1e-3} & \cellcolor[HTML]{FFFFFF}32 & \multicolumn{1}{c|}{\cellcolor[HTML]{FFFFFF}1e-2} & 32 & 2 & 3 \\ \hline
\begin{tabular}[c]{@{}c@{}}FigLang\\ Reddit\end{tabular} & \multicolumn{1}{c|}{\cellcolor[HTML]{FFFFFF}1e-4} & 64 & \multicolumn{1}{c|}{\cellcolor[HTML]{FFFFFF}1e-2} & 32 & 2 & 2 \\ \hline
\end{tabular}%
\caption{Data-specific hyperparameters for \modelName. \textit{lr}: learning rate, \textit{max sents}: Max. sentences considered in a source post while training, \textit{max coms}: Max. comments on post considered while training.}
\end{wraptable}

\textbf{Data-specific baselines.} We experiment with three different baselines per dataset. We compare the model performance using F1 Score, Precision and Recall. These models are known to have reported representative results on the benchmark datasets. For {\textit{Fake news detection}}, (v) \textbf{TCNN-URG} \cite{qian2018neural} utilises a CNN-based network for encoding news content, and a variation auto-encoder (VAE) for modelling the user comments (vi) \textbf{CSI} \cite{ruchansky2017csi} is a hybrid deep learning model that utilizes subtle clues from text, responses, and source post, while modelling the news representation using an LSTM-based network, for fake news detection, and lastly (vii) \textbf{HPA-BLSTM} \cite{guo2018rumor} learns news representations through a word-level, post-level, and event-level user engagements on social media. These turned out to have representative performance for fake news detection and therefore, we consider them as baselines for Politifact, Gossipcop, and AntiVax datasets (which are related to the task of fake news detection). In addition to CSI and HPA-BLSTM as baselines for {\em Hate speech detection} on the HASOC dataset, we use  \textbf{CRNN} as a baseline due to the ability of CNNs to capture the sequential correlation in text. In \textit{rumour detection}, for Twitter15 and Twitter16 datasets, (v) \textbf{AARD} \cite{song2021adversary} uses a weighted-edge transformer-graph network and position-aware adversarial response generator to capture the malicious user attacks while spreading rumours, (vi) \textbf{GCAN} \cite{lu2020gcan} employs a dual co-attention mechanism between source social media post and the underlying propagation patterns, and (vii) \textbf{BiGCN} \cite{chen2021bigcn}, utilizes the original graph structure information and the latent correlation between features assisted by bidirectional-filtering. Further, for Pheme dataset we use (v) \textbf{RumourGAN} \cite{ma2019detect} which adheres to a GAN-based approach, where the generator is designed to produce uncertain or conflicting opinions (voices), complicating the original conversational threads in order to penalise the discriminator to learn better, (vi) \textbf{DDGCN} \cite{korban2020ddgcn}, which models spatial and temporal features of human actions
from their skeletal representations, and (vii) \textbf{STS-NN} \cite{huang2020deep}. On RumourEval, (v) \textbf{DeClarE} \cite{ye2021end} provides a strong baseline. Moreover simple yet effective models like (vi) \textbf{CNN} and (vi) \textbf{MTL-LSTM} show comparable performance, and hence are included in our set of baselines. For the task of (d) \textit{Sarcasm detection} on Figlang (Twitter) and Figlang (Reddit) datasets, we use (v) \textbf{CNN} + \textbf{LSTM} \cite{jain2020sarcasm}, (vi) an ensemble of CNN, LSTM, SVM and MLP \cite{lemmens-etal-2020-sarcasm}, and lastly (vii) \textbf{C-Net} \cite{kumar-jena-etal-2020-c} for efficient sarcasm classification.

\section{Explainability} \label{appendix:explainability}

\textbf{Data annotation.} To evaluate the efficacy of \modelName\ in producing suitable explanations, we fact-check and annotate the Politifact dataset on a sentence-level. Each sentence has the following possible labels -- \textit{true}, \textit{false}, \textit{quote}, \textit{unverified}, \textit{non\_check\_worthy} or \textit{noise}. The annotators were further supposed to arrange the fact-checked sentences in the order of their check-worthiness. We take the help of four expert annotators in the age-group of 25-30 years. The final labels for a sentence were decided on the basis of majority voting amongst the four annotators. To decide the final rank-list (since different annotators might have different opinions about the level of check-worthiness of the sentences), the fourth annotator compiled the final rank-list by referring to the fact-checked rank-lists by the first three annotators using Kendall's $\tau$ and Spearman's $\rho$ rank correlation coefficients, and manually observing the similarities between the three rank-lists. The compiled list is then cross-checked and re-evaluated by the first three annotators for consistency.

\begin{figure}[!t]
\centering
\includegraphics[scale = 0.80]{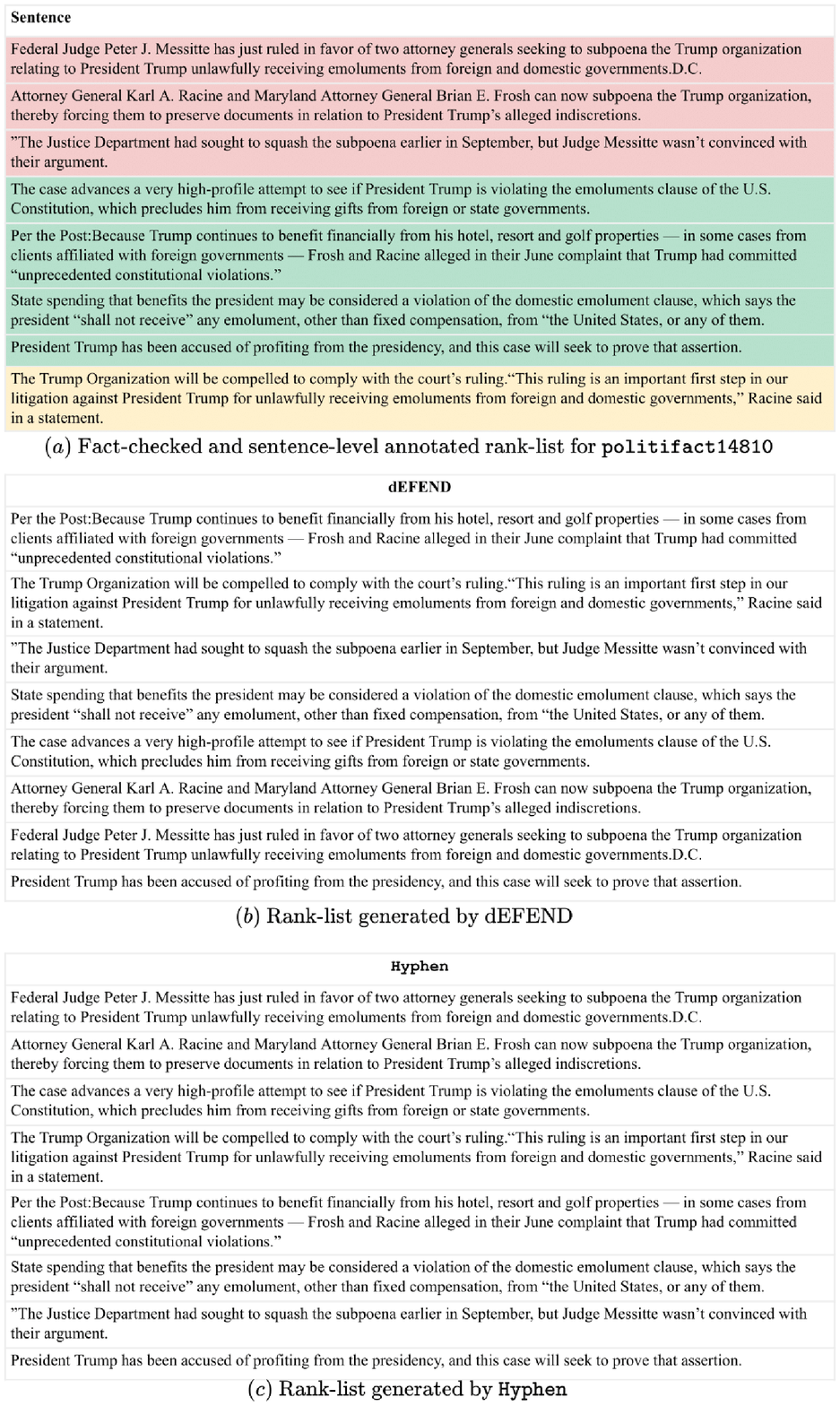}
\caption{Sample rank-lists generated by \modelName-\texttt{hyperbolic} and dEFEND. (a) Ground-truth annotation for \texttt{politifact14810} sample. \textit{Red}: fake sentences, \textit{Green}: true sentences, and \textit{Yellow}: quote. It can be observed that there is almost no correlation between the dEFEND rank-list and the ground-truth. The rank-list produced by \modelName\ is observably quite similar to the annotated list.}
\label{fig:rank_list}
\end{figure}

\textbf{Explainability evaluation.} To evaluate the performance of \modelName\ against the annotated rank-list, we measure the rank-correlation between the two. If \modelName\ predicts a news article in Politifact to be fake, we filter the sentences in the ground-truth annotation with the label \textit{fake} (in the order of their check-worthiness). We adopt a similar procedure in case \modelName\ predicts a news article to be true. This is done because if a news article is fake, we aim to identify the sentences in the article which are misinformation and thus most relevant to the final prediction. Finally, we compare the filtered ground-truth rank-list with the rank-list produced by \modelName\ using Kendall's $\tau$ and Spearman's $\rho$ coefficients. Figure \ref{fig:rank_list} shows sample rank-lists produced by \modelName-\texttt{hyperbolic} and dEFEND \cite{shu2019defend}.


%% file: main.bbl
\begin{thebibliography}{10}

\bibitem{unsvaag2018effects}
Elise~Fehn Unsv{\aa}g and Bj{\"o}rn Gamb{\"a}ck.
\newblock The effects of user features on twitter hate speech detection.
\newblock In {\em Proceedings of the 2nd workshop on abusive language online
  (ALW2)}, pages 75--85, 2018.

\bibitem{shu2019role}
Kai Shu, Xinyi Zhou, Suhang Wang, Reza Zafarani, and Huan Liu.
\newblock The role of user profiles for fake news detection.
\newblock In {\em Proceedings of the 2019 IEEE/ACM international conference on
  advances in social networks analysis and mining}, pages 436--439, 2019.

\bibitem{malhotra2012studying}
Anshu Malhotra, Luam Totti, Wagner Meira~Jr, Ponnurangam Kumaraguru, and
  Virgilio Almeida.
\newblock Studying user footprints in different online social networks.
\newblock In {\em 2012 IEEE/ACM International Conference on Advances in Social
  Networks Analysis and Mining}, pages 1065--1070. IEEE, 2012.

\bibitem{nurrahmi2018indonesian}
Hani Nurrahmi and Dade Nurjanah.
\newblock Indonesian twitter cyberbullying detection using text classification
  and user credibility.
\newblock In {\em 2018 International Conference on Information and
  Communications Technology (ICOIACT)}, pages 543--548. IEEE, 2018.

\bibitem{yang2021rumor}
Xiaoyu Yang, Yuefei Lyu, Tian Tian, Yifei Liu, Yudong Liu, and Xi~Zhang.
\newblock Rumor detection on social media with graph structured adversarial
  learning.
\newblock In {\em Proceedings of the Twenty-Ninth International Conference on
  International Joint Conferences on Artificial Intelligence}, pages
  1417--1423, 2021.

\bibitem{guille2013predicting}
Adrien Guille, Hakim Hacid, and C{\'e}cile Favre.
\newblock Predicting the temporal dynamics of information diffusion in social
  networks.
\newblock {\em arXiv preprint arXiv:1302.5235}, 2013.

\bibitem{pitas2016graph}
Ioannis Pitas.
\newblock {\em Graph-based social media analysis}, volume~39.
\newblock CRC Press, 2016.

\bibitem{guille2013information}
Adrien Guille, Hakim Hacid, Cecile Favre, and Djamel~A Zighed.
\newblock Information diffusion in online social networks: A survey.
\newblock {\em ACM Sigmod Record}, 42(2):17--28, 2013.

\bibitem{lu2020gcan}
Yi-Ju Lu and Cheng-Te Li.
\newblock Gcan: Graph-aware co-attention networks for explainable fake news
  detection on social media.
\newblock {\em arXiv preprint arXiv:2004.11648}, 2020.

\bibitem{shu2019beyond}
Kai Shu, Suhang Wang, and Huan Liu.
\newblock Beyond news contents: The role of social context for fake news
  detection.
\newblock In {\em Proceedings of the twelfth ACM international conference on
  web search and data mining}, pages 312--320, 2019.

\bibitem{nguyen2020fang}
Van-Hoang Nguyen, Kazunari Sugiyama, Preslav Nakov, and Min-Yen Kan.
\newblock Fang: Leveraging social context for fake news detection using graph
  representation.
\newblock In {\em Proceedings of the 29th ACM international conference on
  information \& knowledge management}, pages 1165--1174, 2020.

\bibitem{zubiaga2017exploiting}
Arkaitz Zubiaga, Maria Liakata, and Rob Procter.
\newblock Exploiting context for rumour detection in social media.
\newblock In {\em International conference on social informatics}, pages
  109--123. Springer, 2017.

\bibitem{gao2017detecting}
Lei Gao and Ruihong Huang.
\newblock Detecting online hate speech using context aware models.
\newblock {\em arXiv preprint arXiv:1710.07395}, 2017.

\bibitem{banarescu2012abstract}
Laura Banarescu, Claire Bonial, Shu Cai, Madalina Georgescu, Kira Griffitt, Ulf
  Hermjakob, Kevin Knight, Philipp Koehn, Martha Palmer, and Nathan Schneider.
\newblock Abstract meaning representation (amr) 1.0 specification.
\newblock In {\em Parsing on Freebase from Question-Answer Pairs. In
  Proceedings of the 2013 Conference on Empirical Methods in Natural Language
  Processing. Seattle: ACL}, pages 1533--1544, 2012.

\bibitem{flanigan2014discriminative}
Jeffrey Flanigan, Sam Thomson, Jaime~G Carbonell, Chris Dyer, and Noah~A Smith.
\newblock A discriminative graph-based parser for the abstract meaning
  representation.
\newblock In {\em Proceedings of the 52nd Annual Meeting of the Association for
  Computational Linguistics (Volume 1: Long Papers)}, pages 1426--1436, 2014.

\bibitem{https://doi.org/10.48550/arxiv.2110.07855}
Peiyi Wang, Liang Chen, Tianyu Liu, Damai Dai, Yunbo Cao, Baobao Chang, and
  Zhifang Sui.
\newblock Hierarchical curriculum learning for amr parsing, 2021.

\bibitem{liu2018matching}
Bang Liu, Ting Zhang, Fred~X Han, Di~Niu, Kunfeng Lai, and Yu~Xu.
\newblock Matching natural language sentences with hierarchical sentence
  factorization.
\newblock In {\em Proceedings of the 2018 World Wide Web Conference}, pages
  1237--1246, 2018.

\bibitem{cannon1997hyperbolic}
James~W Cannon, William~J Floyd, Richard Kenyon, Walter~R Parry, et~al.
\newblock Hyperbolic geometry.
\newblock {\em Flavors of geometry}, 31(59-115):2, 1997.

\bibitem{cooley1965algorithm}
James~W Cooley and John~W Tukey.
\newblock An algorithm for the machine calculation of complex fourier series.
\newblock {\em Mathematics of computation}, 19(90):297--301, 1965.

\bibitem{https://doi.org/10.48550/arxiv.2112.12809}
Maunika Tamire, Srinivas Anumasa, and P.~K. Srijith.
\newblock Bi-directional recurrent neural ordinary differential equations for
  social media text classification, 2021.

\bibitem{fei2015social}
Geli Fei and Bing Liu.
\newblock Social media text classification under negative covariate shift.
\newblock In {\em Proceedings of the 2015 Conference on Empirical Methods in
  Natural Language Processing}, pages 2347--2356, 2015.

\bibitem{liu2019roberta}
Yinhan Liu, Myle Ott, Naman Goyal, Jingfei Du, Mandar Joshi, Danqi Chen, Omer
  Levy, Mike Lewis, Luke Zettlemoyer, and Veselin Stoyanov.
\newblock Roberta: A robustly optimized bert pretraining approach.
\newblock {\em arXiv preprint arXiv:1907.11692}, 2019.

\bibitem{nguyen2020bertweet}
Dat~Quoc Nguyen, Thanh Vu, and Anh~Tuan Nguyen.
\newblock Bertweet: A pre-trained language model for english tweets.
\newblock {\em arXiv preprint arXiv:2005.10200}, 2020.

\bibitem{alsentzer2019publicly}
Emily Alsentzer, John~R Murphy, Willie Boag, Wei-Hung Weng, Di~Jin, Tristan
  Naumann, and Matthew McDermott.
\newblock Publicly available clinical bert embeddings.
\newblock {\em arXiv preprint arXiv:1904.03323}, 2019.

\bibitem{guo-etal-2020-benchmarking}
Yuting Guo, Xiangjue Dong, Mohammed~Ali Al-Garadi, Abeed Sarker, Cecile Paris,
  and Diego~Moll{\'a} Aliod.
\newblock Benchmarking of transformer-based pre-trained models on social media
  text classification datasets.
\newblock In {\em Proceedings of the The 18th Annual Workshop of the
  Australasian Language Technology Association}, pages 86--91, Virtual
  Workshop, December 2020. Australasian Language Technology Association.

\bibitem{https://doi.org/10.48550/arxiv.2105.03824}
James Lee-Thorp, Joshua Ainslie, Ilya Eckstein, and Santiago Ontanon.
\newblock Fnet: Mixing tokens with fourier transforms, 2021.

\bibitem{qian2018neural}
Feng Qian, Chengyue Gong, Karishma Sharma, and Yan Liu.
\newblock Neural user response generator: Fake news detection with collective
  user intelligence.
\newblock In {\em IJCAI}, volume~18, pages 3834--3840, 2018.

\bibitem{ruchansky2017csi}
Natali Ruchansky, Sungyong Seo, and Yan Liu.
\newblock Csi: A hybrid deep model for fake news detection.
\newblock In {\em Proceedings of the 2017 ACM on Conference on Information and
  Knowledge Management}, pages 797--806, 2017.

\bibitem{zubiaga2018discourse}
Arkaitz Zubiaga, Elena Kochkina, Maria Liakata, Rob Procter, Michal Lukasik,
  Kalina Bontcheva, Trevor Cohn, and Isabelle Augenstein.
\newblock Discourse-aware rumour stance classification in social media using
  sequential classifiers.
\newblock {\em Information Processing \& Management}, 54(2):273--290, 2018.

\bibitem{lee2018discourse}
Kangwook Lee, Sanggyu Han, and Sung-Hyon Myaeng.
\newblock A discourse-aware neural network-based text model for document-level
  text classification.
\newblock {\em Journal of Information Science}, 44(6):715--735, 2018.

\bibitem{https://doi.org/10.48550/arxiv.1805.06413}
Devamanyu Hazarika, Soujanya Poria, Sruthi Gorantla, Erik Cambria, Roger
  Zimmermann, and Rada Mihalcea.
\newblock Cascade: Contextual sarcasm detection in online discussion forums,
  2018.

\bibitem{https://doi.org/10.48550/arxiv.1607.00976}
Silvio Amir, Byron~C. Wallace, Hao Lyu, and Paula Carvalho Mário~J. Silva.
\newblock Modelling context with user embeddings for sarcasm detection in
  social media, 2016.

\bibitem{shu2019defend}
Kai Shu, Limeng Cui, Suhang Wang, Dongwon Lee, and Huan Liu.
\newblock defend: Explainable fake news detection.
\newblock In {\em Proceedings of the 25th ACM SIGKDD international conference
  on knowledge discovery \& data mining}, pages 395--405, 2019.

\bibitem{chami2019hyperbolic}
Ines Chami, Zhitao Ying, Christopher R{\'e}, and Jure Leskovec.
\newblock Hyperbolic graph convolutional neural networks.
\newblock {\em Advances in neural information processing systems}, 32, 2019.

\bibitem{https://doi.org/10.48550/arxiv.1912.03046}
Yiding Zhang, Xiao Wang, Xunqiang Jiang, Chuan Shi, and Yanfang Ye.
\newblock Hyperbolic graph attention network, 2019.

\bibitem{dai2021hyperbolic}
Jindou Dai, Yuwei Wu, Zhi Gao, and Yunde Jia.
\newblock A hyperbolic-to-hyperbolic graph convolutional network.
\newblock In {\em Proceedings of the IEEE/CVF Conference on Computer Vision and
  Pattern Recognition}, pages 154--163, 2021.

\bibitem{https://doi.org/10.48550/arxiv.2010.12135}
Shichao Zhu, Shirui Pan, Chuan Zhou, Jia Wu, Yanan Cao, and Bin Wang.
\newblock Graph geometry interaction learning, 2020.

\bibitem{frigo2005design}
Matteo Frigo and Steven~G Johnson.
\newblock The design and implementation of fftw3.
\newblock {\em Proceedings of the IEEE}, 93(2):216--231, 2005.

\bibitem{https://doi.org/10.48550/arxiv.2111.13587}
John Guibas, Morteza Mardani, Zongyi Li, Andrew Tao, Anima Anandkumar, and
  Bryan Catanzaro.
\newblock Adaptive fourier neural operators: Efficient token mixers for
  transformers, 2021.

\bibitem{https://doi.org/10.48550/arxiv.1910.12933}
Ines Chami, Rex Ying, Christopher Ré, and Jure Leskovec.
\newblock Hyperbolic graph convolutional neural networks, 2019.

\bibitem{zhang2021hype}
Chengkun Zhang and Junbin Gao.
\newblock Hype-han: Hyperbolic hierarchical attention network for semantic
  embedding.
\newblock In {\em Proceedings of the Twenty-Ninth International Conference on
  International Joint Conferences on Artificial Intelligence}, pages
  3990--3996, 2021.

\bibitem{lu2016hierarchical}
Jiasen Lu, Jianwei Yang, Dhruv Batra, and Devi Parikh.
\newblock Hierarchical question-image co-attention for visual question
  answering.
\newblock {\em Advances in neural information processing systems}, 29, 2016.

\bibitem{shu2020fakenewsnet}
Kai Shu, Deepak Mahudeswaran, Suhang Wang, Dongwon Lee, and Huan Liu.
\newblock Fakenewsnet: A data repository with news content, social context, and
  spatiotemporal information for studying fake news on social media.
\newblock {\em Big data}, 8(3):171--188, 2020.

\bibitem{hayawi2022anti}
Kadhim Hayawi, Sakib Shahriar, Mohamed~Adel Serhani, Ikbal Taleb, and
  Sujith~Samuel Mathew.
\newblock Anti-vax: a novel twitter dataset for covid-19 vaccine misinformation
  detection.
\newblock {\em Public health}, 203:23--30, 2022.

\bibitem{mandl2019overview}
Thomas Mandl, Sandip Modha, Prasenjit Majumder, Daksh Patel, Mohana Dave,
  Chintak Mandlia, and Aditya Patel.
\newblock Overview of the hasoc track at fire 2019: Hate speech and offensive
  content identification in indo-european languages.
\newblock In {\em Proceedings of the 11th forum for information retrieval
  evaluation}, pages 14--17, 2019.

\bibitem{8118443}
Cody Buntain and Jennifer Golbeck.
\newblock Automatically identifying fake news in popular twitter threads.
\newblock In {\em 2017 IEEE International Conference on Smart Cloud
  (SmartCloud)}, pages 208--215, 2017.

\bibitem{ma-etal-2018-rumor}
Jing Ma, Wei Gao, and Kam-Fai Wong.
\newblock Rumor detection on {T}witter with tree-structured recursive neural
  networks.
\newblock In {\em Proceedings of the 56th Annual Meeting of the Association for
  Computational Linguistics (Volume 1: Long Papers)}, pages 1980--1989,
  Melbourne, Australia, July 2018. Association for Computational Linguistics.

\bibitem{gorrell2019semeval}
Genevieve Gorrell, Elena Kochkina, Maria Liakata, Ahmet Aker, Arkaitz Zubiaga,
  Kalina Bontcheva, and Leon Derczynski.
\newblock Semeval-2019 task 7: Rumoureval, determining rumour veracity and
  support for rumours.
\newblock In {\em Proceedings of the 13th International Workshop on Semantic
  Evaluation}, pages 845--854, 2019.

\bibitem{ghosh2020report}
Debanjan Ghosh, Avijit Vajpayee, and Smaranda Muresan.
\newblock A report on the 2020 sarcasm detection shared task.
\newblock {\em arXiv preprint arXiv:2005.05814}, 2020.

\bibitem{guo2018rumor}
Han Guo, Juan Cao, Yazi Zhang, Junbo Guo, and Jintao Li.
\newblock Rumor detection with hierarchical social attention network.
\newblock In {\em Proceedings of the 27th ACM international conference on
  information and knowledge management}, pages 943--951, 2018.

\bibitem{chen2021bigcn}
Zhixian Chen, Tengfei Ma, Zhihua Jin, Yangqiu Song, and Yang Wang.
\newblock Bigcn: A bi-directional low-pass filtering graph neural network.
\newblock {\em arXiv preprint arXiv:2101.05519}, 2021.

\bibitem{song2021adversary}
Yun-Zhu Song, Yi-Syuan Chen, Yi-Ting Chang, Shao-Yu Weng, and Hong-Han Shuai.
\newblock Adversary-aware rumor detection.
\newblock In {\em Findings of the Association for Computational Linguistics:
  ACL-IJCNLP 2021}, pages 1371--1382, 2021.

\bibitem{korban2020ddgcn}
Matthew Korban and Xin Li.
\newblock Ddgcn: A dynamic directed graph convolutional network for action
  recognition.
\newblock In {\em European Conference on Computer Vision}, pages 761--776.
  Springer, 2020.

\bibitem{ma2019detect}
Jing Ma, Wei Gao, and Kam-Fai Wong.
\newblock Detect rumors on twitter by promoting information campaigns with
  generative adversarial learning.
\newblock In {\em The world wide web conference}, pages 3049--3055, 2019.

\bibitem{huang2020deep}
Qi~Huang, Chuan Zhou, Jia Wu, Luchen Liu, and Bin Wang.
\newblock Deep spatial--temporal structure learning for rumor detection on
  twitter.
\newblock {\em Neural Computing and Applications}, pages 1--11, 2020.

\bibitem{ye2021end}
Aoshuang Ye, Lina Wang, Run Wang, Wenqi Wang, Jianpeng Ke, and Danlei Wang.
\newblock An end-to-end rumor detection model based on feature aggregation.
\newblock {\em Complexity}, 2021, 2021.

\bibitem{9498905}
Mostafa Karimzadeh, Samuel~Martin Schwegler, Zhongliang Zhao, Torsten Braun,
  and Susana Sargento.
\newblock Mtl-lstm: Multi-task learning-based lstm for urban traffic flow
  forecasting.
\newblock In {\em 2021 International Wireless Communications and Mobile
  Computing (IWCMC)}, pages 564--569, 2021.

\bibitem{jain2020sarcasm}
Deepak Jain, Akshi Kumar, and Geetanjali Garg.
\newblock Sarcasm detection in mash-up language using soft-attention based
  bi-directional lstm and feature-rich cnn.
\newblock {\em Applied Soft Computing}, 91:106198, 2020.

\bibitem{lemmens-etal-2020-sarcasm}
Jens Lemmens, Ben Burtenshaw, Ehsan Lotfi, Ilia Markov, and Walter Daelemans.
\newblock Sarcasm detection using an ensemble approach.
\newblock In {\em Proceedings of the Second Workshop on Figurative Language
  Processing}, pages 264--269, Online, July 2020. Association for Computational
  Linguistics.

\bibitem{kumar-jena-etal-2020-c}
Amit Kumar~Jena, Aman Sinha, and Rohit Agarwal.
\newblock {C}-net: Contextual network for sarcasm detection.
\newblock In {\em Proceedings of the Second Workshop on Figurative Language
  Processing}, pages 61--66, Online, July 2020. Association for Computational
  Linguistics.

\bibitem{https://doi.org/10.48550/arxiv.2005.02819}
Max Kochurov, Rasul Karimov, and Serge Kozlukov.
\newblock Geoopt: Riemannian optimization in pytorch, 2020.

\bibitem{gu2018learning}
Albert Gu, Frederic Sala, Beliz Gunel, and Christopher R{\'e}.
\newblock Learning mixed-curvature representations in product spaces.
\newblock In {\em International Conference on Learning Representations}, 2018.

\end{thebibliography}
